\documentclass[journal]{IEEEtran}
\usepackage{amssymb}
\usepackage{amsmath}
\usepackage{graphicx}
\usepackage{flushend}
\usepackage{booktabs}

\usepackage[T1]{fontenc} 
\usepackage{multirow} 
\usepackage{xcolor} 
\usepackage[caption=false,font=footnotesize]{subfig}

\newcommand{\modelname}{MARLIN}

\newcommand{\intraclasslossname}{Frequency-Aware Prototype Compactness}
\newcommand{\interclasslossname}{Pairwise Frequency-Weighted Margin Separation}

\ifCLASSINFOpdf
\else
\fi
\begin{document}
%

\title{MARLIN: Multiscale And Directional Sparse Representation Learning for Imbalanced SAR Ship Classification}

%
%
%

\author{Benyamin~Hosseiny,
        Kamirul~Kamirul,
        Odysseas~Pappas,
        and~Alin~Achim,~\IEEEmembership{Senior~Member,~IEEE}
\thanks{The authors  are with the VI-Labs, University of
Bristol, BS1 6AZ Bristol, U.K.  (e-mail: ben.hosseiny@bristol.ac.uk; kamirul.kamirul@bristol.ac.uk;
o.pappas@bristol.ac.uk; alin.achim@bristol.ac.uk)}
\thanks{ This work was supported in part by the Engineering and Physical Sciences Research Council under Grant EP/X525674/1 EPSRC Impact Acceleration Account - University of Bristol 2022 and in part by the Dstl under the DASA Defence Rapid Impact Open Call. The work of Kamirul Kamirul was supported by Lembaga Pengelola Dana Pendidikan (LPDP), Ministry of Finance of the Republic of Indonesia. }}
    
%
%

\markboth{Journal of \LaTeX\ Class Files,~Vol.~14, No.~8, November~2026}%
{Shell \MakeLowercase{\textit{et al.}}: Bare Demo of IEEEtran.cls for IEEE Journals}
%



\maketitle

\begin{abstract}
Despite recent progress in Synthetic Aperture Radar (SAR) maritime monitoring, accurate ship classification remains challenging due to limited labeled data, severe class imbalance, large intra-class variability, fine-grained inter-class similarity, and variations in target scale, orientation, and backscattering structure. 
To address these challenges, this paper proposes \modelname{}, a class-frequency-aware framework with directional sparse and hierarchical multi-scale feature extraction for ship type classification in SAR images. 
The proposed architecture first constructs two complementary representations of each input image, where each views are processed by parallel encoder branches. Each branch begins with a directional sparse convolutional stem designed to capture elongated and orientation-dependent ship scattering patterns, followed by customized hierarchical multi-scale feature extraction blocks.
To improve learning under imbalanced class distributions, we further introduce a class-frequency-aware metric learning objective. The proposed objective combines margin-based classification with class-frequency-weighted prototype compactness and class-pair frequency weighted margin separation. This formulation explicitly shapes the embedding space by encouraging samples from the same class to form compact clusters while enforcing stronger separation for class pairs involving underrepresented ship categories.
Extensive experiments on three complementary SAR ship-classification datasets: OpenSARShip, FUSAR-Ship, and NASTaR demonstrate that the proposed framework consistently outperforms baseline and advanced comparison models across different datasets and classification scenarios. The results show that in addition to being lightweight and parameter efficient, \modelname{} improves both overall accuracy and class-balanced metrics, confirming its effectiveness for robust SAR ship-type classification under limited and imbalanced training conditions. The codes are available at https://github.com/benyaminhosseiny/marlin.
\end{abstract}

\begin{IEEEkeywords}
Maritime Monitoring, Target Recognition, Deep Learning, Metric Learning
\end{IEEEkeywords}

%
\IEEEpeerreviewmaketitle

\section{Introduction}

Synthetic Aperture Radar (SAR) is widely used in maritime monitoring due to its all-weather, day-and-night imaging capability. In addition to detecting vessels, SAR-based classification can identify vessel types, enabling applications such as traffic analysis, fisheries enforcement, border security, search and rescue, and anomaly detection \cite{Qiao2025,Awais2026}.

Ship detection in SAR is facilitated by the strong radar return that ships produce compared to the sea surface background, making them appear as bright targets in imagery. Traditional approaches to ship detection have relied on threshold-based detectors such as CFAR \cite{Cyprien2024}, often utilizing statistical models of sea clutter, including truncated statistics \cite{Tao2016} or superpixel architectures \cite{Pappas2018}, while more modern deep learning approaches try to learn the relations between ship and local background directly from the data  \cite {Qiao2025,Kamirul2025,Li2022}. 

Subsequent to ship detection comes the task of ship classification, i.e. the determination of vessel type, a characteristic crucial to many applications such as traffic analysis and illegal fishing monitoring. 
This classification task is fundamentally challenging as the detected targets often exhibit little discernible radiometric, textural or shape features \cite{Awais2026}. 
It can be further challenging due to both large intra-class variability and fine-grained class similarity. Ships of the same type, e.g. fishing vessels, can greatly vary in size and shape while at the same time, classes such as ``cargo'' and ``container ship'' may greatly overlap on those same features.

Deep learning has become the dominant approach for SAR ship classification. However, a common thread across early approaches was that simple direct re-training of deep neural backbones did not necessarily lead to optimal performance, leading to research into more complex architectures that can learn more meaningful representations from the targets and their local background and/or incorporate prior knowledge and handcrafted features, while also addressing issues around limited labeled data and class imbalance. 
Multi-scale features, attention mechanism, and adaptive feature selection have become increasingly important due to the variety and complexities in the shapes of each ship type.
For instance, a polarimetric information fusion model with geometric feature embeddings was proposed in \cite{Zhang2022}.
Multiscale feature attention with an adaptive-weighted classifier has been proposed to enhance discriminative components at different feature scales in \cite{Wang2023MultiscaleAttention}. 
Deformable and multi-scale feature learning has also been explored in \cite{Chen2024DCNMSFF}, which integrates deformable convolution into the backbone and uses self-attentive feature fusion within a feature hierarchy to improve SAR ship classification \cite{Chen2024DCNMSFF}. 
These studies demonstrate the importance of multiscale and adaptive feature extraction for distinguishing visually similar ship categories.

Another challenge in SAR ship classification is data scarcity and class imbalance.
Publicly available datasets including OpenSARShip \cite{Li2017}, FUSAR-Ship \cite{Hou2020}, and NASTaR \cite{Hosseiny2026} are either small or exhibit significant class imbalance, with certain vessel types heavily overrepresented. This makes classification more challenging and often necessitates imbalance-aware training. 
Attempts to mitigate these challenges include synthetic data generation, and novel learning techniques other than supervised learning, such as metric learning, unsupervised learning, or self-supervised learning \cite{rizaev2022modeling,Rizaev2022,Hosseiny2023,Zhang2024}.
A number of SAR ship image simulators exist in the literature, including analytical methods such as AssenSAR
~\cite{rizaev2022modeling,Rizaev2022} and that of \cite{Guo2018}, as well as more modern generative approaches such as the denoising diffusion model-based Ship-Go approach \cite{Zhang2024}, and Text-to-image generation using a Tokens-to-Token Vision Transformer backbone \cite{Wang2024TextToImageSE}.
To mitigate the effect of class imbalance, \cite{Shao2018} utilised a weighted loss function dependent on class distribution. 
In another work, \cite{Awais2025} explored curriculum learning, and loss function weighting strategies have been explored to address the extreme imbalances of FUSAR-Ship. 

More recently, increasing attention has been directed toward developing generalized SAR ship classification frameworks that can operate effectively under limited labeled data through self-supervised learning and few-shot adaptation.
For example, \cite{Zhang2023} proposed a few-shot SAR ship classifier based on scattering-point topology and a multi-branch embedding network with a bi-classifier.
\cite{li2024predicting} proposed a self-supervised multiscale representation learning framework for generalized automatic SAR target recognition, including ship types, achieving strong performance under few-shot fine-tuning conditions. Similarly, \cite{li2025saratr} developed a SAR target recognition foundation model, and demonstrated that self-supervised pretraining can improve downstream SAR target classification.

Nevertheless, despite the significant progress achieved in recent years, SAR ship classification remains challenging, as various methods attempt to address them, often in part and with varying degrees of success. 
Typical spatial convolutions used in conventional CNNs are not well-suited to the oriented, elongated ship targets. The use of a single-scale image can greatly inhibit the methods ability to make use of fine-grained information while also utilizing broader background information. 
Furthermore, due to dataset class imbalance, conventional sampling and loss formulations are often dominated by well-represented classes and may produce poorly structured embeddings for minority categories.
These limitations motivate the development of a unified framework that captures directional ship structure and scale-dependent scattering patterns, combines complementary target-focused and contextual views, and explicitly accounts for class imbalance during training.

In this paper, we present a novel method for ship classification that addresses the above challenges in a unified framework. \modelname{} is designed to take into account the visual characteristics of ship targets, with their elongated shape and highly varied size and backscattering structures. 
It constructs complementary central-crop and pooled-context representations of each input image, allowing target-specific structural information and broader contextual cues to be learned in parallel. 
Directional sparse convolutions are used to capture elongated and sparsely distributed ship responses, while hierarchical deformable and dilated feature extraction enables adaptation to variations in target scale, shape and scattering structure. 
The resulting embedding features are optimized using an imbalance-aware metric-learning objective, designed to improve class compactness and separation without allowing majority classes to dominate the learned feature space.

The main contributions of this study are summarized as follows: 
\begin{itemize}

\item We propose \modelname{}, a novel SAR ship classification framework that combines directional sparse convolution and hierarchical multiscale feature extraction to capture ship geometry, scale variations, and scattering characteristics.

\item We introduce a class-frequency-aware metric-learning objective that improves the representation and discrimination of underrepresented ship categories under imbalanced data distributions.

\item We evaluate the proposed framework on three publicly available SAR ship classification datasets, OpenSARShip, FUSAR-Ship, and NASTaR, demonstrating consistent performance improvements across different sensors, resolutions, and imaging conditions.

\end{itemize}

The remainder of this paper is structured as follows. 
Section~\ref{sec:Methodology} presents the proposed \modelname{} framework and provides a detailed description of its main components. 
Section~\ref{sec:Experiments} describes the experimental setup, including the datasets, preprocessing procedures, evaluation metrics, ablation studies, comparison protocols, and discussion of the results. 
Finally, Section~\ref{sec:Conclusion} provides concluding remarks.

\section{Methodology}
\label{sec:Methodology}
\subsection{Overview}

We propose a deep learning framework for ship-type classification under imbalanced training conditions. The framework is designed by considering the discriminative visual characteristics of different ship types, which are often elongated along specific directions and exhibit substantial variations in size, shape, and backscattering structure. The proposed model, referred to as \modelname{}, integrates dual-view input preparation, directional sparse convolution, hierarchical multi-scale feature extraction, dual-branch feature fusion, and imbalance-aware metric learning.

\begin{figure*}[!t]
\centering
\includegraphics[width=1\textwidth]{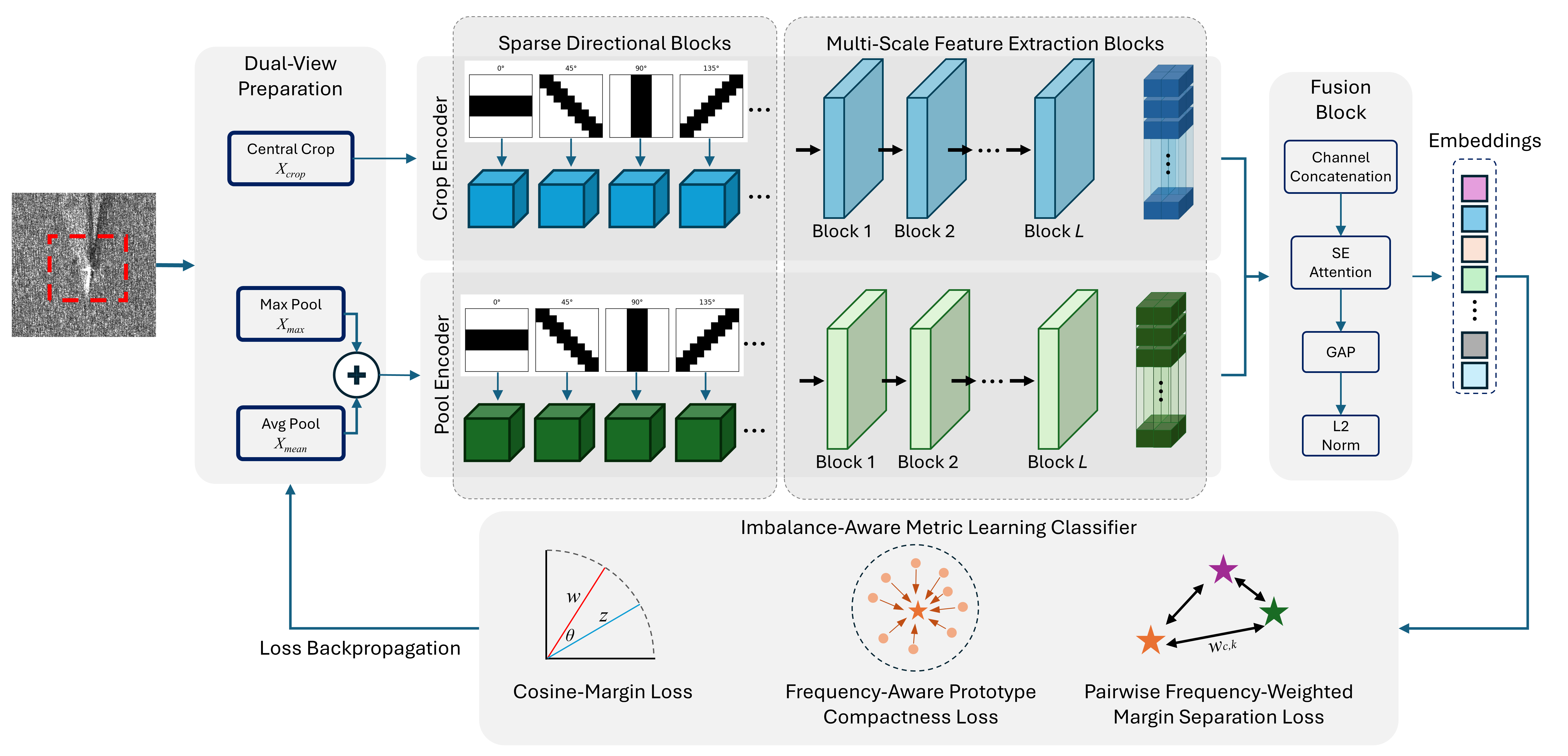}
\caption{Overview of the proposed \modelname{} for ship type classification}\label{fig:framework}
\end{figure*}

The overall architecture, shown in Fig.~\ref{fig:framework}, first decomposes an input image into two complementary representations: a central crop and a pooled contextual view. These views are processed by parallel encoder branches, each consisting of a directional sparse convolutional stem followed by stacked multi-scale deformable and dilated feature extraction blocks. This dual-branch design enables the model to jointly capture target-focused spatial details and broader contextual information. The resulting features are fused and transformed into a normalized embedding, which is optimized using an imbalance-aware metric-learning loss to enhance class compactness and inter-class separability while improving classification performance under imbalanced training conditions.

\subsection{Dual-View Pooling}

This module is designed to enhance information extraction while guiding the model toward the region where the target is likely to be located. Motivated by \cite{zhang2021}, the input image is compressed into smaller spatial representations reducing the influence of local noise while highlighting strong backscattering responses associated with the target structure.

Given an input image $X \in \mathbb{R}^{C_{\mathrm{in}} \times H \times W}$, where $C_{\mathrm{in}}$ denotes the number of input channels, and $H$ and $W$ denote the height and width of the input image, respectively, the module generates three intermediate representations with identical spatial dimensions: a central crop ($X_{\mathrm{crop}}$), a max-pooled view ($X_{\max}$), and an average-pooled view ($X_{\mathrm{mean}}$).
The target spatial size is determined by a reduction factor $r$ as $H' = \left\lfloor \frac{H}{r} \right\rfloor$ and $W' = \left\lfloor \frac{W}{r} \right\rfloor $, where $H'$ and $W'$ denote the height and width of the generated views, respectively.

The three representations are computed as
\begin{equation}
\begin{aligned}
X_{\mathrm{crop}}
&=
\mathrm{CenterCrop}_{H' \times W'}(X), \\
X_{\max}
&=
\mathrm{AdaptiveMaxPool}_{H' \times W'}(X), \\
X_{\mathrm{mean}}
&=
\mathrm{AdaptiveAvgPool}_{H' \times W'}(X).
\end{aligned}
\end{equation}

The max-pooled and average-pooled representations are then combined to form a pooled contextual view:
\begin{equation}
X_{\mathrm{pool}}
=
\frac{1}{2}
\left(
X_{\max}
+
X_{\mathrm{mean}}
\right).
\end{equation}

Here, $X_{\mathrm{pool}}$ denotes the pooled contextual view obtained by averaging the max-pooled and average-pooled representations.
Therefore, the network receives two complementary views:
the central crop preserves target-focused spatial details, whereas the pooled contextual view combines the strong-response preservation of max pooling with the smoother and less noise-sensitive statistics of average pooling.

\subsection{Directional Sparse Convolution}

Ships are elongated objects with strong orientation-dependent structures. To exploit this property, each branch begins with a Directional Sparse Convolutional Stem. This module applies convolutional filters constrained to a set of predefined orientations.

The set of considered directions is defined as $\Theta=\{\theta_1, \theta_2, \ldots, \theta_{|\Theta|}\}$, where, $\Theta$ denotes the set of predefined directional orientations, $\theta_i$ denotes the $i$-th orientation angle, and $|\Theta|$ denotes the number of considered orientations.

For a directional convolution kernel of size $k_{\mathrm{dir}} \times k_{\mathrm{dir}}$, a binary mask is generated for each direction $\theta$. Let $(x,y)$ denote coordinates relative to the kernel center. The distance from $(x,y)$ to the line with orientation $\theta$ is defined as
\begin{equation}
d_{\theta}(x,y)
=
\left|
\sin(\theta)x
-
\cos(\theta)y
\right|.
\end{equation}
Here, $k_{\mathrm{dir}}$ denotes the directional convolution kernel size, $(x,y)$ denotes a coordinate location relative to the kernel center, $\theta$ denotes the considered orientation angle, and $d_{\theta}(x,y)$ denotes the perpendicular distance from $(x,y)$ to the line associated with orientation $\theta$.

The directional mask is defined as
\begin{equation}
M_{\theta}(x,y)
=
\begin{cases}
1, & d_{\theta}(x,y) \leq \frac{\tau}{2},\\
0, & \mathrm{otherwise},
\end{cases}
\end{equation}
where $M_{\theta}(x,y)$ denotes the binary mask value at location $(x,y)$ for direction $\theta$, and $\tau$ denotes the mask thickness controlling the width of the active directional support region.

The effective convolution kernel for orientation $\theta$ is obtained through element-wise multiplication:
\begin{equation}
\widetilde{W}_{\theta}
=
W_{\theta}
\odot
M_{\theta}.
\end{equation}
Here, $W_{\theta}$ denotes the learnable convolution kernel associated with direction $\theta$, $\widetilde{W}_{\theta}$ denotes the effective masked directional kernel, and $\odot$ denotes element-wise multiplication. Only the active masked positions contribute to the directional response, making the operation sparse and parameter-efficient. Therefore, the directional convolution response is computed as $Y_{\theta}=X * \widetilde{W}_{\theta}$, where $Y_{\theta}$ denotes the directional response for orientation $\theta$, and $*$ denotes the convolution operator.

To retain unconstrained local patterns, the directional branch is combined with a standard convolution branch. A residual path is also included to improve gradient flow. Therefore, the output of the Directional Sparse Convolutional Stem is computed as
\begin{equation}
\begin{aligned}
Y_{\mathrm{ds}}
&=
\mathrm{Concat}
\left[
Y_{\theta_1},
Y_{\theta_2},
\ldots,
Y_{\theta_{|\Theta|}}
\right], \\
Y_{\mathrm{std}}
&=
\mathrm{Conv}_{k_s \times k_s}(X), \\
Y_{\mathrm{ds\text{-}out}}
&=
\mathrm{Concat}
\left[
Y_{\mathrm{ds}},
Y_{\mathrm{std}}
\right], \\
Y_{\mathrm{res\text{-}ds}}
&=
Y_{\mathrm{ds\text{-}out}}
+
\mathcal{P}(X).
\end{aligned}
\end{equation}
Here, $Y_{\mathrm{ds}}$ denotes the concatenated directional response, $Y_{\mathrm{std}}$ denotes the output of the standard convolution branch, $k_s$ denotes the kernel size of the standard convolution, $Y_{\mathrm{ds\text{-}out}}$ denotes the concatenated output of the directional and standard convolution branches, $Y_{\mathrm{res\text{-}ds}}$ denotes the residual output of the directional sparse convolutional stem, and $\mathcal{P}(\cdot)$ denotes either an identity mapping or a projection used to match feature dimensions.

\begin{figure}[!t]
\centering

\subfloat[Orientation masks\label{fig:dsconv_masks}]{
    \includegraphics[width=0.8\columnwidth]{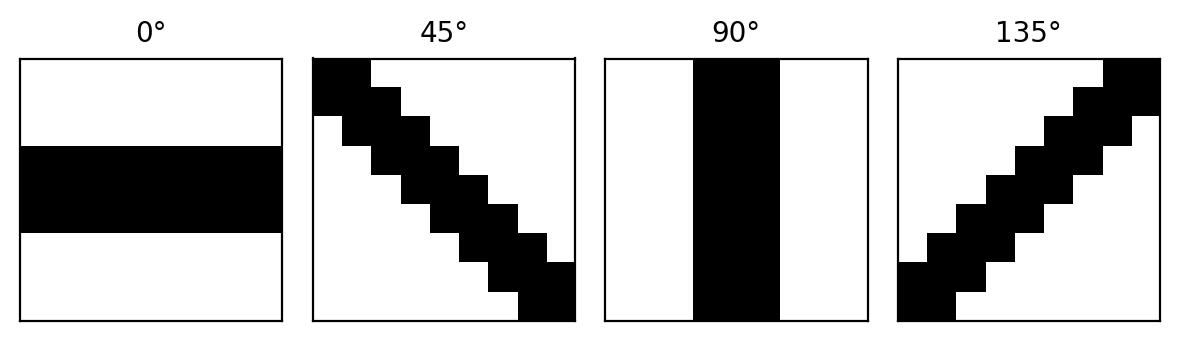}
}

\vspace{0.6em}

\subfloat[Directional sparse kernels\label{fig:dsconv_kernels}]{
    \includegraphics[width=0.8\columnwidth]{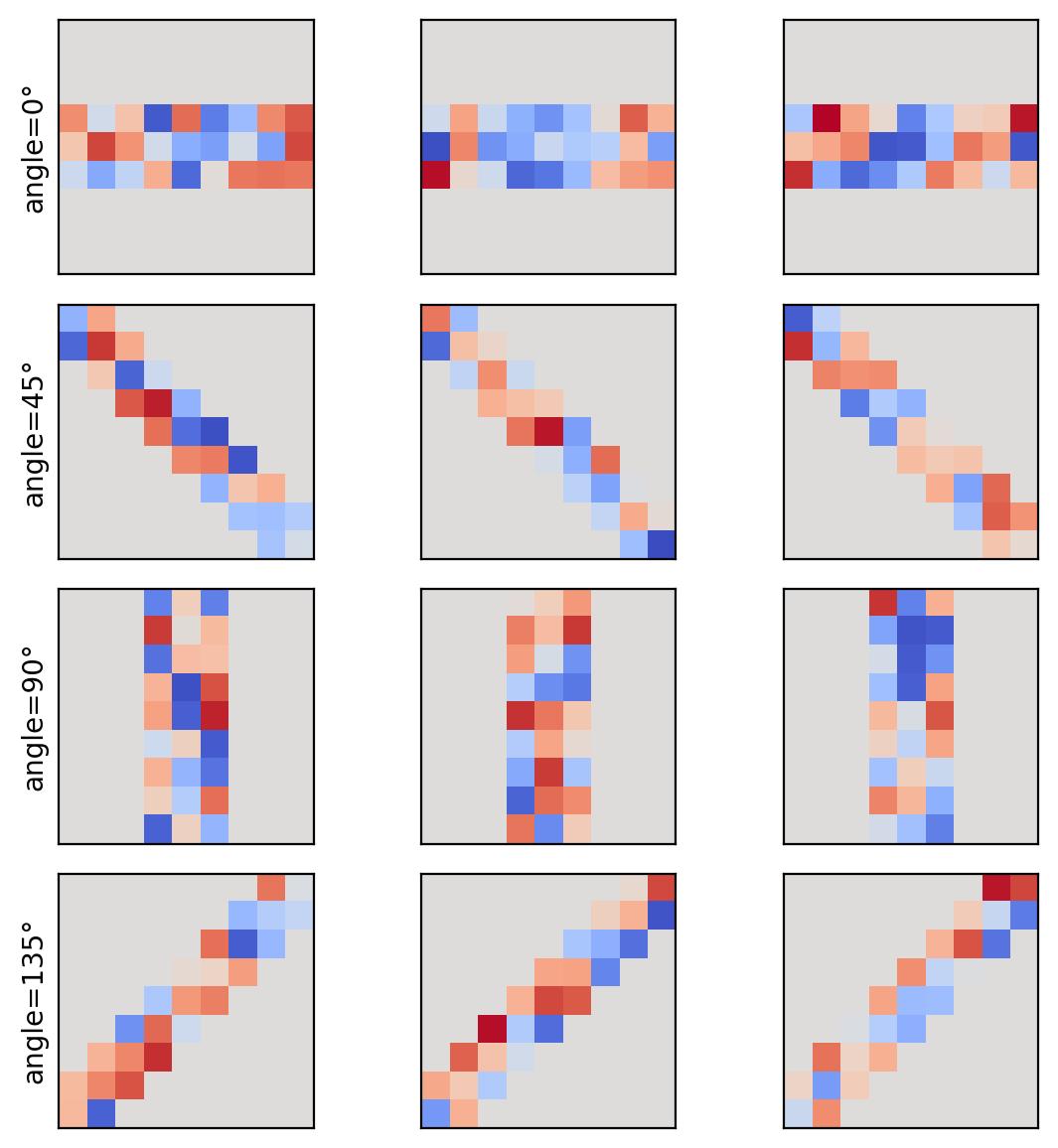}
}

\caption{Visualization of the Directional Sparse Convolutional Stem. The upper subfigure shows the fixed orientation masks, and the lower subfigure shows the corresponding directional sparse kernels.}
\label{fig:dsconv_visualization}

\end{figure}

Figure~\ref{fig:dsconv_visualization} illustrates an example of the proposed orientation masks and directional sparse kernels. In this example, four orientations are considered, $\Theta=\{0^\circ,45^\circ,90^\circ,135^\circ\}$, and $q=3$ filters are assigned to each orientation, resulting in $|\Theta|\times q=4\times3=12$ directional output channels. 

\subsection{Hierarchical Multi-Scale Feature Extraction}

To capture the significant variability in ship scale, shape, and orientation, we integrate hierarchical multi-scale feature extraction with residual mixed-pooling fusion within a unified design. Ships often appear as elongated structures with diverse sizes and irregular geometries, making standard convolutions with fixed receptive fields insufficient.

After the directional sparse convolutional stem, each branch is processed by a sequence of Multi-Scale Deformable Feature Extraction Blocks. Each block contains parallel dilated and deformable convolutional branches designed to capture local, contextual, and geometrically adaptive structures at different scales. Each convolutional branch includes batch normalization and a non-linear activation function.

For the $\ell$-th block, the multi-scale branch outputs are computed as
\begin{equation}
\begin{aligned}
F_{d}^{(\ell)}
&=
\phi
\left(
\mathrm{BN}
\left(
\mathrm{DilConv}_{k_d \times k_d}^{d}
\left(
F^{(\ell-1)}
\right)
\right)
\right), \\
F_{\mathrm{def}}^{(\ell)}
&=
\phi
\left(
\mathrm{BN}
\left(
\mathrm{DefConv}_{k_{\mathrm{def}} \times k_{\mathrm{def}}}^{d_{\mathrm{def}}}
\left(
F^{(\ell-1)}
\right)
\right)
\right), \\
\mathcal{F}^{(\ell)}
&=
\left\{
F_d^{(\ell)} \mid d \in \mathcal{D}
\right\}
\cup
\left\{
F_{\mathrm{def}}^{(\ell)}
\right\}.
\end{aligned}
\end{equation}
Here, $\ell$ denotes the block index, $F^{(\ell-1)}$ denotes the input feature map to the $\ell$-th block, $F_d^{(\ell)}$ denotes the output of the dilated convolution branch with dilation rate $d$, $\mathcal{D}$ denotes the set of dilation rates, $k_d$ denotes the kernel size of the dilated convolution, $F_{\mathrm{def}}^{(\ell)}$ denotes the output of the deformable convolution branch, $k_{\mathrm{def}}$ denotes the deformable convolution kernel size, $d_{\mathrm{def}}$ denotes the deformable convolution dilation rate, $\phi(\cdot)$ denotes the activation function, $\mathrm{BN}(\cdot)$ denotes batch normalization, and $\mathcal{F}^{(\ell)}$ denotes the set of branch outputs in the $\ell$-th multi-scale block.

The outputs of the parallel branches are fused using a Residual Mixed-Pooling Fusion Block. The branch outputs are first concatenated along the channel dimension. A pointwise projection is then applied to fuse the concatenated multi-scale features and control the output channel dimensionality. A residual shortcut is added to improve information flow, and a mixed average--max pooling operator is finally applied for spatial downsampling:
\begin{equation}
\begin{aligned}
F_{\mathrm{cat}}^{(\ell)}
&=
\mathrm{Concat}
\left[
\mathcal{F}^{(\ell)}
\right], \\
F_{\mathrm{proj}}^{(\ell)}
&=
\phi
\left(
\mathrm{BN}
\left(
\mathrm{Conv}_{1 \times 1}
\left(
F_{\mathrm{cat}}^{(\ell)}
\right)
\right)
\right), \\
F_{\mathrm{skip}}^{(\ell)}
&=
\mathcal{S}
\left(
F^{(\ell-1)}
\right), \\
F_{\mathrm{res}}^{(\ell)}
&=
F_{\mathrm{proj}}^{(\ell)}
+
F_{\mathrm{skip}}^{(\ell)}, \\
F^{(\ell)}
&=
\frac{1}{2}
\left[
\mathrm{AvgPool}
\left(
F_{\mathrm{res}}^{(\ell)}
\right)
+
\mathrm{MaxPool}
\left(
F_{\mathrm{res}}^{(\ell)}
\right)
\right].
\end{aligned}
\end{equation}

Here, $F_{\mathrm{cat}}^{(\ell)}$ denotes the concatenated multi-scale feature representation, $F_{\mathrm{proj}}^{(\ell)}$ denotes the projected representation obtained after applying a $1\times1$ convolution to the concatenated multi-scale features, $F_{\mathrm{skip}}^{(\ell)}$ denotes the shortcut representation obtained from the block input, and $F_{\mathrm{res}}^{(\ell)}$ denotes the residual fused representation before spatial downsampling. The shortcut function $\mathcal{S}(\cdot)$ is implemented as an identity mapping when the input feature map $F^{(\ell-1)}$ has the same channel dimension as $F_{\mathrm{proj}}^{(\ell)}$; otherwise, $\mathcal{S}(\cdot)$ is implemented using a $1\times1$ convolution to match the channel dimensions required for residual addition. The final output of the block is denoted by $F^{(\ell)}$.

\begin{figure*}[!t]
\centering
\includegraphics[width=0.9\textwidth]{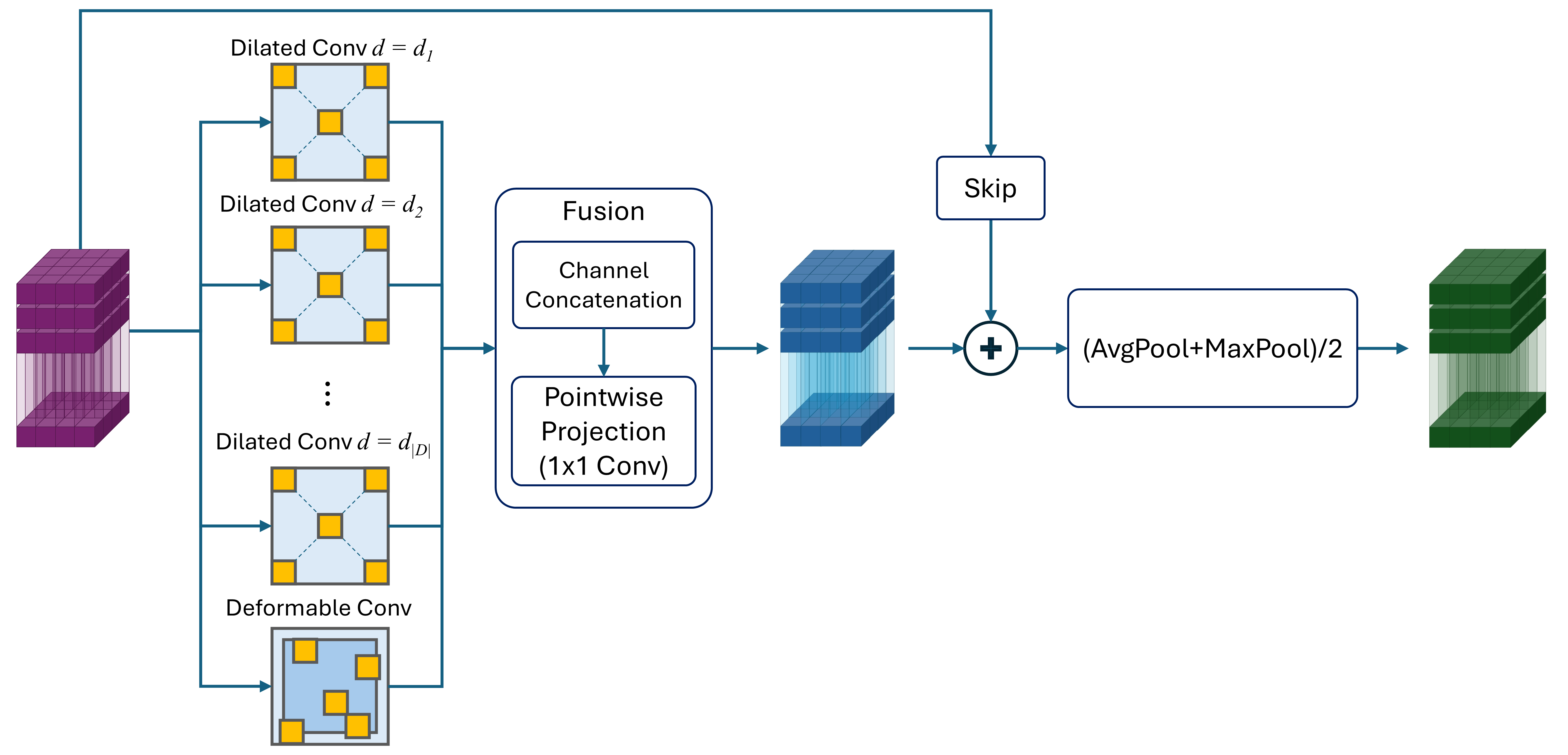}
\caption{Structure of the proposed hierarchical multi-scale feature extraction block. The input feature map $F^{(\ell-1)}$ is processed by parallel dilated convolution branches with different dilation rates $d \in \mathcal{D}$ and a deformable convolution branch.
The branch outputs are concatenated along the channel dimension to form $F_{\mathrm{cat}}^{(\ell)}$, followed by a $1\times1$ pointwise projection to obtain $F_{\mathrm{proj}}^{(\ell)}$. After adding a residual shortcut from the input features, the resulting feature map $F_{\mathrm{res}}^{(\ell)}$ is downsampled using mixed average--max pooling, producing the block output $F^{(\ell)}$ for the next stage.}\label{fig:msfe}
\end{figure*}

This fusion strategy simultaneously increases feature abstraction, reduces spatial resolution, and lowers the computational cost of subsequent layers. The average-pooling component contributes stable contextual smoothing, while the max-pooling component preserves strong discriminative activations. The structure of the proposed hierarchical multi-scale feature extraction block is illustrated in Fig.~\ref{fig:msfe}.

\subsection{Encoder Overview}

As described in the previous sections, the proposed encoder consists of two independent branches, where each branch is composed of a Directional Sparse Convolutional Stem followed by a stack of Multi-Scale Deformable Feature Extraction Blocks. The two branches process complementary views of the same input image: the central crop view and the pooled contextual view. The central crop branch focuses on target-centered spatial details, whereas the pooled branch captures a compressed contextual representation that combines salient responses and averaged structural information.

Let $X_{\mathrm{crop}}$ and $X_{\mathrm{pool}}$ denote the two prepared views. The corresponding branch features are obtained as $F_{\mathrm{crop}} = E_{\mathrm{crop}}(X_{\mathrm{crop}})$ and $F_{\mathrm{pool}} = E_{\mathrm{pool}}(X_{\mathrm{pool}})$, where the functions $E_{\mathrm{crop}}(\cdot)$ and $E_{\mathrm{pool}}(\cdot)$ denote the corresponding encoder branches, which share the same architectural design but have independent learnable parameters.

Each branch encoder can be expressed as
\begin{equation}
E_b(\cdot)
=
B_L
\circ
B_{L-1}
\circ
\cdots
\circ
B_1
\circ
S_{\mathrm{ds}}(\cdot),
\qquad
b \in \{\mathrm{crop},\mathrm{pool}\}.
\end{equation}
where, $b$ denotes the branch index, $S_{\mathrm{ds}}(\cdot)$ denotes the Directional Sparse Convolutional Stem, $B_{\ell}$ denotes the $\ell$-th Multi-Scale Deformable Feature Extraction Block, $L$ denotes the number of stacked multi-scale blocks, and $\circ$ denotes function composition.

After hierarchical feature extraction, the concatenated dual-view feature representation is obtained by fusing the outputs of the two branches along the channel dimension as $F_{\mathrm{view}}=\mathrm{Concat}[F_{\mathrm{crop}},F_{\mathrm{pool}}]$.

To adaptively emphasize informative channels in the fused representation, a squeeze-and-excitation attention operation is applied, yielding $F_{\mathrm{att}}=\mathrm{SE}(F_{\mathrm{view}})$.
The attended fused representation is then converted into a compact feature vector using global average pooling, i.e., $u=\mathrm{GAP}(F_{\mathrm{att}})$.
Finally, the embedding is $\ell_2$-normalized before being passed to the metric-learning classifieras $z=\frac{u}{\|u\|_2}$, where, $z$ denotes the normalized embedding vector and $\|u\|_2$ denotes the Euclidean norm of $u$.

\subsection{Loss Function}

To address the imbalance in the training samples, we introduce a class-frequency aware prototype-margin metric-learning objective. The main idea is to encourage samples from the same ship type to form compact clusters while enforcing sufficient separation between different ship types. Unlike standard cross-entropy, which mainly optimizes decision boundaries, the proposed objective directly regularizes the geometry of the learned feature space. This is particularly important for minority ship classes, which may otherwise be overwhelmed by majority classes.

The proposed loss contains three complementary components. First, a margin-based normalized classification term based on additive cosine-margin softmax improves angular discrimination between ship classes. Second, a class-frequency weighted prototype compactness term pulls each embedding toward the prototype of its ground-truth class. Third, a pairwise frequency-weighted margin separation term pushes embeddings from different classes apart when their distance is smaller than a required class-pair margin. 

Given the final encoder output, global average pooling and $\ell_2$ normalization are applied to obtain the embedding vector $z_i$ for sample $i$. The overall objective is written as
\begin{equation}
\mathcal{L}
=
\mathcal{L}_{\mathrm{cls}}
+
\lambda_{\mathrm{intra}}
\mathcal{L}_{\mathrm{intra}}
+
\lambda_{\mathrm{inter}}
\mathcal{L}_{\mathrm{inter}}
\end{equation}
where, $\mathcal{L}$ denotes the overall loss, $\mathcal{L}_{\mathrm{cls}}$ denotes the cosine-margin classification loss, $\mathcal{L}_{\mathrm{intra}}$ denotes the class-frequency weighted prototype compactness loss, and $\mathcal{L}_{\mathrm{inter}}$ denotes the pairwise frequency-weighted margin separation loss. The coefficients $\lambda_{\mathrm{intra}}$ and $\lambda_{\mathrm{inter}}$ control the relative contributions of the compactness and separation terms, respectively.

\subsubsection{Cosine-Margin Loss}

The classification component is based on additive cosine-margin softmax, also known as CosFace \cite{wang2018}. Standard cross-entropy optimizes class probabilities but does not explicitly enforce a margin between classes in the embedding space. This limitation can be problematic for visually similar ship types, especially when some classes have very few samples.

To address this, both the embedding vector $z_i$ and the classifier weight vector $w_c$ are $\ell_2$-normalized, and their similarity is computed using cosine similarity:
\begin{equation}
\cos\theta_{i,c}
=
z_i^\top w_c .
\end{equation}
Here, $z_i$ denotes the normalized embedding vector of sample $i$, $w_c$ denotes the normalized classifier weight vector for class $c$, and $\theta_{i,c}$ denotes the angle between $z_i$ and $w_c$.

For the ground-truth class $y_i$, an additive margin is subtracted from the cosine similarity:
\begin{equation}
\tilde{\ell}_{i,c}
=
\begin{cases}
s(\cos\theta_{i,c} - m), & c = y_i,\\
s\cos\theta_{i,c}, & c \neq y_i,
\end{cases}
\end{equation}
where $\tilde{\ell}_{i,c}$ denotes the modified logit of sample $i$ for class $c$, $y_i$ denotes the ground-truth class label of sample $i$, $m$ denotes the additive cosine margin, and $s$ denotes the logit scaling factor. 
Here, the role of $m$ is to make the classification criterion stricter for the correct class. By subtracting $m$ from the cosine similarity of the ground-truth class, the model is required to produce a higher original similarity $\cos\theta_{i,y_i}$ in order for the correct class to dominate the softmax output. Equivalently, the embedding $z_i$ must be more closely aligned with its corresponding class weight $w_{y_i}$, resulting in a smaller angle $\theta_{i,y_i}$. This encourages stronger intra-class compactness and larger angular separation between different classes, leading to a more discriminative embedding space.

The classification loss is then computed using cross-entropy:
\begin{equation}
\mathcal{L}_{\mathrm{cls}}
=
-
\frac{1}{B}
\sum_{i=1}^{B}
\log
\frac{
\exp(\tilde{\ell}_{i,y_i})
}{
\sum_{c=1}^{N_c}
\exp(\tilde{\ell}_{i,c})
},
\end{equation}
where, $B$ denotes the mini-batch size, $N_c$ denotes the number of ship classes, and $c$ indexes the classes.

This term makes the classification criterion stricter than ordinary cross-entropy by requiring the correct class to be separated from competing classes by an additional cosine margin. Therefore, it encourages more discriminative angular embeddings.

\subsubsection{\intraclasslossname{}}

To encourage samples from the same ship type to cluster together, each class is represented by a prototype vector $\mu_c$, which acts as the class center in the embedding space. The \intraclasslossname{} loss is defined as
\begin{equation}
\mathcal{L}_{\mathrm{intra}}
=
\frac{1}{B}
\sum_{i=1}^{B}
w_{y_i}^{\mathrm{intra}}
\left\|
z_i - \mu_{y_i}
\right\|_2^2 .
\end{equation}
where, $\mu_{y_i}$ denotes the prototype vector of the ground-truth class $y_i$, and $w_{y_i}^{\mathrm{intra}}$ denotes a class-frequency-dependent intra-class compactness weight.
The intra-class weight is defined as

\begin{equation}
w_c^{\mathrm{intra}}
=
\frac{n_c}{\bar{n}}.
\end{equation}
where, $n_c$ denotes the number of training samples in class $c$, $n_j$ denotes the number of training samples in class $j$, $\bar{n}$ is the average number of training samples per class.

The weight $w_c^{\mathrm{intra}}$ controls the strength of the compactness constraint for class $c$. Since this weight multiplies the squared distance between an embedding and its class prototype, a larger value of $w_c^{\mathrm{intra}}$ produces a stronger penalty when samples from class $c$ are far from their prototype. Consequently, embeddings from classes with larger weights are pulled more strongly toward their corresponding class centers.

Therefore, this loss term pulls each embedding toward the prototype of its own class and helps obtain compact class clusters. By incorporating class frequency information, the compactness constraint can reduce the tendency of majority classes to occupy overly broad regions of the embedding space.

\subsubsection{\interclasslossname{}}

The inter-class term encourages embeddings from different ship types to be sufficiently separated. For sample pairs belonging to different classes, a penalty is applied only when their distance is smaller than a required margin:
\begin{equation}
\mathcal{L}_{\mathrm{inter}}
=
\frac{1}{|\mathcal{P}_{\mathrm{inter}}|}
\sum_{\substack{i<j \\ y_i \neq y_j}}
w_{y_i,y_j}^{\mathrm{inter}}
\left[
M_{y_i,y_j}
-
\left\|
z_i - z_j
\right\|_2
\right]_+^2 .
\end{equation}
where, $\mathcal{P}_{\mathrm{inter}}$ denotes the set of inter-class sample pairs in the mini-batch, $w_{y_i,y_j}^{\mathrm{inter}}$ denotes the inter-class weighting factor for the class pair $(y_i,y_j)$, $M_{y_i,y_j}$ denotes the required minimum separation margin between classes $y_i$ and $y_j$, and $[a]_+=\max(a,0)$ denotes the hinge operator.

To account for class imbalance, the pairwise inter-class weight is computed from the number of training samples in each class:
\begin{equation}
w_{c,k}^{\mathrm{inter}}
=
\left(
\frac{\bar{n}}
{\sqrt{n_c n_k}}
\right)^{\beta},
\qquad c \neq k .
\end{equation}
Here, $w_{c,k}^{\mathrm{inter}}$ denotes the inter-class weight for the class pair $(c,k)$, $n_c$ and $n_k$ denote the numbers of training samples in classes $c$ and $k$, respectively, $\bar{n}$ denotes the mean number of training samples per class, and $\beta$ denotes an imbalance-sensitivity parameter that controls how strongly minority-class pairs are emphasized.

The diagonal entries, which represent the inter-class weight for pairs from the same class, are set to zero in order to be excluded from the inter-class separation loss ($w_{c,c}^{\mathrm{inter}} = 0$).

\begin{figure}[!t]
\centering
\includegraphics[width=0.5\textwidth]{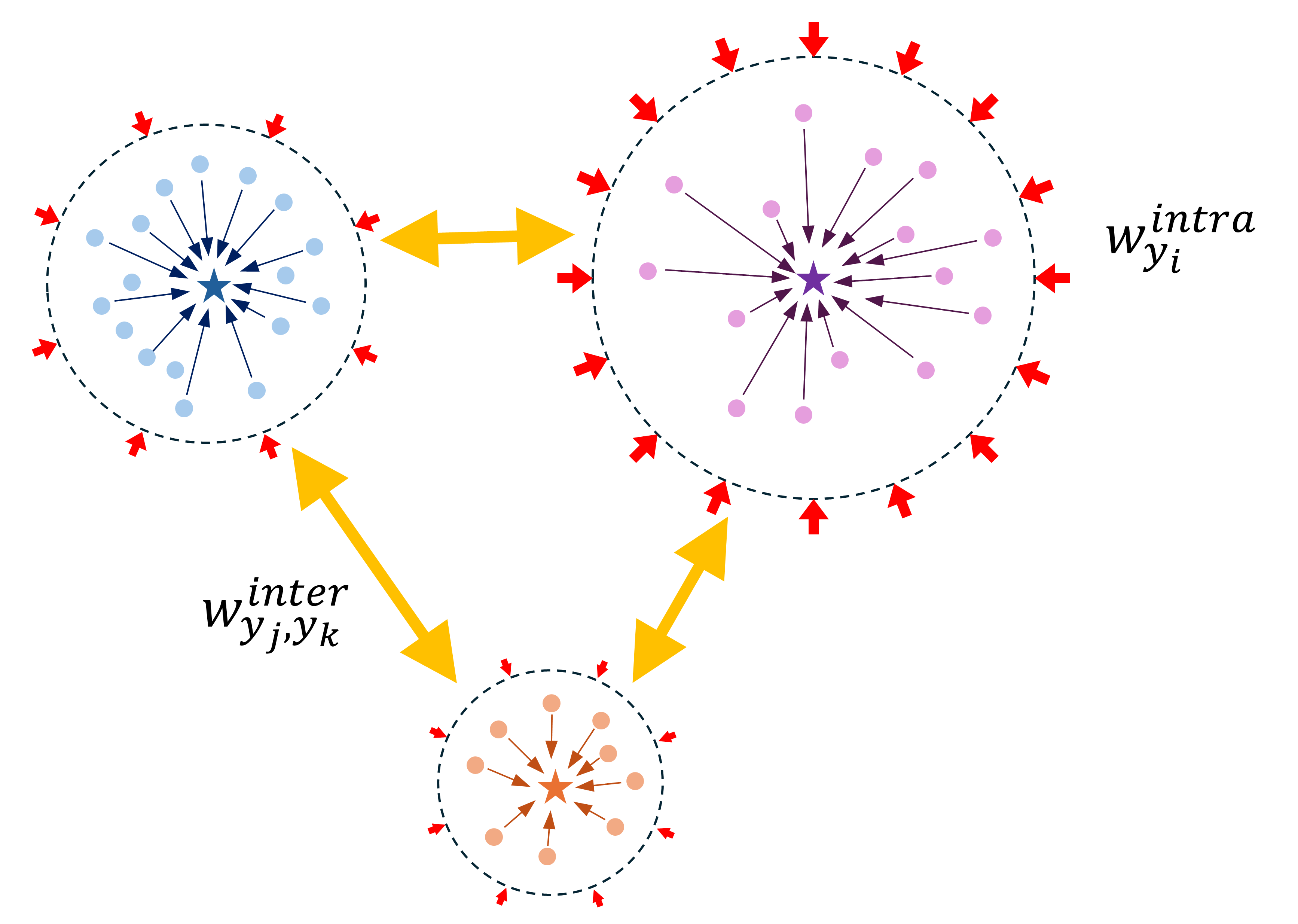}
\caption{
Overview of the proposed intra-class compactness and inter-class separation objectives. In the embedding space, the \intraclasslossname{} term pulls each sample embedding $z_i$ toward its corresponding class prototype $\mu_{y_i}$, with the compactness strength controlled by the class-frequency-dependent weight $w_{y_i}^{\mathrm{intra}}$. The \interclasslossname{} term enforces separation between samples from different classes, where the pairwise inter-class weight $w_{y_i,y_j}^{\mathrm{inter}}$ assigns stronger penalties to class pairs involving underrepresented classes, encouraging better separation of minority ship categories.
}\label{fig:loss}
\end{figure}

Therefore, pairs involving minority classes are assigned stronger penalties and larger separation requirements. This prevents rare ship types from being embedded too close to majority classes and improves minority-class discrimination. Figure~\ref{fig:loss} provides a conceptual illustration of the proposed compactness and separation losses in a two-dimensional embedding space. The intra-class term pulls samples toward their corresponding class prototypes, while the inter-class term pushes different-class samples apart when their distance violates the required margin. The class-frequency-based weights control the strength of these constraints, allowing the loss to account for imbalanced class distributions.

\section{Experiments and Results}
\label{sec:Experiments}
\subsection{Datasets}

To evaluate the developed ship-type classification framework, experiments were conducted on three SAR ship datasets: OpenSARShip \cite{Li2017}, FUSAR-Ship \cite{Hou2020}, and NASTaR \cite{Hosseiny2026}. These datasets provide complementary evaluation conditions in terms of SAR sensor, spatial resolution, ship-type distribution, imaging characteristics, and class imbalance.

\subsubsection{OpenSARShip}

OpenSARShip consists of SAR ship image chips extracted from Sentinel-1 C-band images and annotated with Automatic Identification System (AIS) information and has been widely used as a benchmark for evaluating SAR ship-recognition algorithms \cite{Li2017}. In this study, we used the dual-polarized VV and VH Ground Range Detected (GRD) image patches from OpenSARShip. To ensure a consistent input size, each patch was resampled to $128 \times 128$ pixels.

Two classification scenarios were considered. The four-class scenario includes Bulk Carrier, Container Ship, Fishing, and Tanker. The six-class scenario extends this setting by including Bulk Carrier, Container Ship, Dredger, Fishing, Tanker, and Tug.

\subsubsection{NASTaR}

NASTaR is a NovaSAR-based Automated Ship Target Recognition dataset generated from S-band SAR imagery \cite{Hosseiny2026}. Similar to the other datasets used in this study, it contains AIS-matched ship patches and includes multiple ship categories, with samples separated into inshore and offshore subsets. Compared with OpenSARShip and FUSAR-Ship, NASTaR introduces a different SAR frequency band and acquisition domain, making it useful for evaluating sensor-domain robustness.

In this study, NASTAR's offshore samples were used, and each image patch was center-cropped to $256 \times 256$ pixels. The considered classification setting includes three ship categories: Cargo, Tanker, and Fishing.

\subsubsection{FUSAR-Ship}

FUSAR-Ship is a high-resolution SAR--AIS matched dataset constructed from Gaofen-3 high-resolution C-band SAR imagery \cite{Hou2020}. In this work, each image patch was center-cropped to $256 \times 256$ pixels to obtain a consistent spatial resolution across samples. The considered classification setting includes five ship categories: Bulk Carrier, Container Ship, Fishing, General Cargo, and Tanker.

\subsection{Comparison with Advanced Deep Learning Models}

The proposed model was compared with several advanced deep learning backbones, including ResNet50 \cite{resnet},  EfficientNetV2-M \cite{efficientnetv2}, DenseNet121 \cite{huang2017densely}, DenseNet201 \cite{huang2017densely}, and Vision Transformer (ViT) \cite{vit} models, as well as SAR-JEPA \cite{li2024predicting} and SARATR-X \cite{li2025saratr}, which are specifically designed for SAR target recognition.

For these comparative experiments, the datasets were divided into 70\% training, 10\% validation, and 20\% testing subsets.
The following evaluation metrics were used: Overall Accuracy (OA), Average Accuracy (AA), Average Precision (APr), and Average F1-score (AF1). All results are reported as mean $\pm$ standard deviation across cross-validation splits. 

Additionally, class-wise performance was investigated using three types of confusion matrices. The raw confusion matrix provides an overview of the overall classification outcomes, whereas the row-normalized and column-normalized confusion matrices provide recall- and precision-oriented perspectives, respectively. Specifically, the diagonal elements of the row-normalized matrix correspond to the per-class recall, indicating the proportion of correctly identified samples within each true class. In contrast, the diagonal elements of the column-normalized matrix represent the per-class precision, reflecting the reliability of predictions assigned to each class.

\subsubsection{OpenSARShip Four-Class Scenario}
\label{subsubsec:OpenSARShip_FourClass_Scenario}
\begin{figure*}[!t]
\centering
\includegraphics[width=0.9\textwidth]{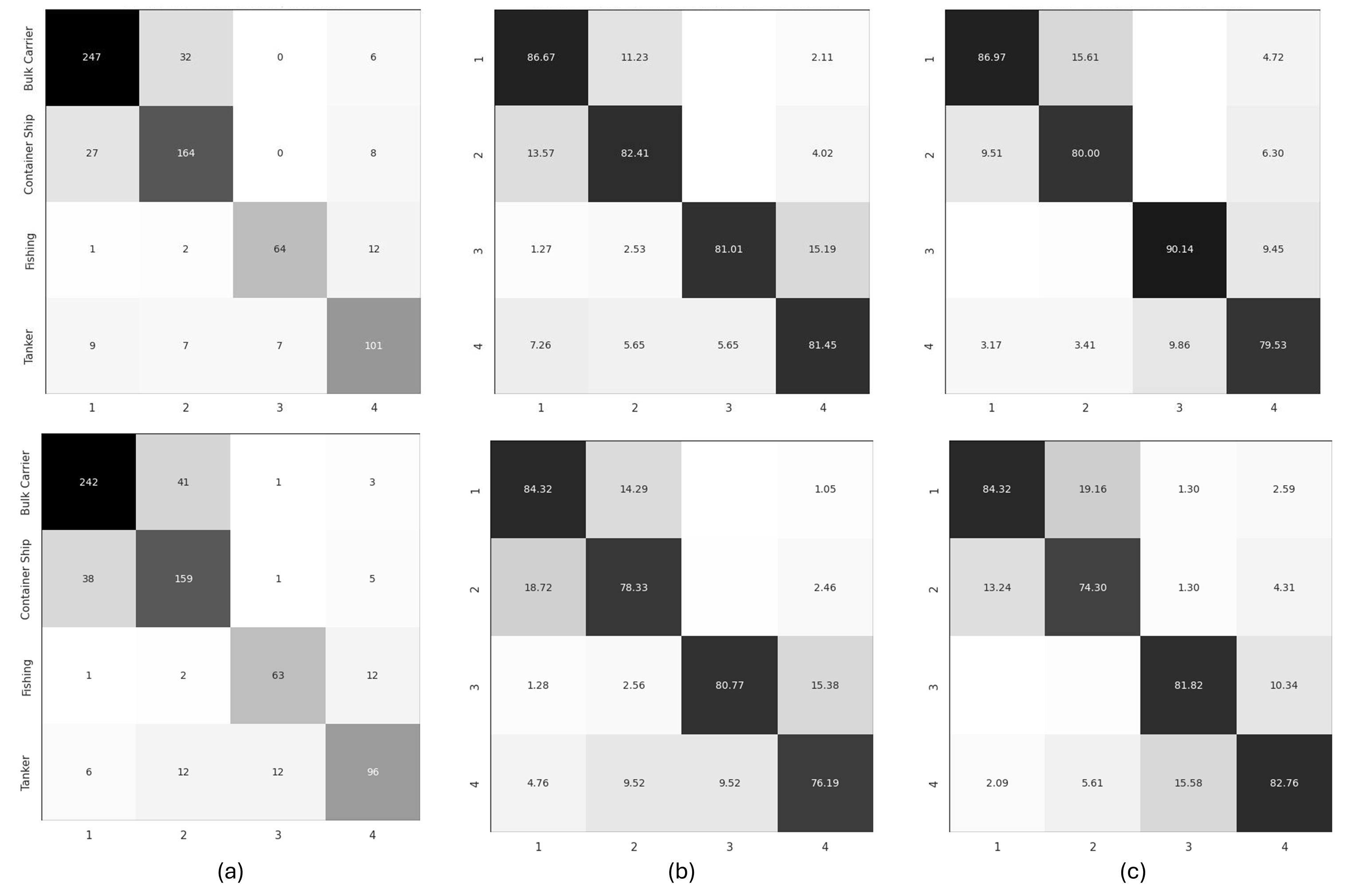}
\caption{Confusion matrices for the OpenSARShip four-class classification experiments. The first row shows the results of the proposed model, while the second row shows the results of the best-performing state-of-the-art model. (a) raw confusion matrix, (b) row-normalized confusion matrix (recall), and (c) column-normalized confusion matrix (precision).}\label{fig:confmat_opensarship4}
\end{figure*}

The OpenSARShip four-class scenario comprises Bulk Carrier, Container Ship, Fishing, and Tanker classes. The quantitative comparison presented in Table~\ref{tab:opensarship_4class} demonstrates that the proposed method consistently outperforms the competing approaches, achieving the highest scores across all evaluation metrics. These results indicate the effectiveness of the proposed framework in learning discriminative representations for maritime target classification under diverse operating conditions.
To further investigate the class-wise performance, 
Fig.~\ref{fig:confmat_opensarship4} compares the confusion matrices of the proposed model and the best-performing state-of-the-art model.
It can be observed that the proposed method exhibits stronger diagonal responses across almost all matrix representations, indicating higher class-wise accuracy and reduced inter-class confusion. 
The row-normalized confusion matrices reveal consistently higher recall values for all vessel categories, demonstrating the model's ability to correctly identify samples belonging to each class.
Similarly, the column-normalized confusion matrices show improved precision, indicating fewer false-positive predictions and higher confidence in the assigned class labels. 
Furthermore, the off-diagonal elements are generally less pronounced compared with those of the state-of-the-art method, suggesting a substantial reduction in class confusion and misclassification errors.

\begin{table}[!t] 
\footnotesize
\caption{Performance comparison on the OpenSARShip four-class classification dataset. The best and second-best results are shown in bold and underlined, respectively.} 
\label{tab:opensarship_4class} 
\centering \footnotesize \setlength{\tabcolsep}{3pt} 
\begin{tabular}{lcccc} 
\toprule 
Model & OA (\%) & AA (\%) & APr (\%) & AF1 (\%) \\ 
\midrule 
ResNet50 \cite{resnet} & $76.7 \pm 2.0$ & $75.5 \pm 2.1$ & $77.2 \pm 1.5$ & $75.8 \pm 2.2$ \\ 
EffiNetV2-M \cite{efficientnetv2} & $77.2 \pm 2.8$ & $76.1 \pm 2.7$ & $78.0 \pm 1.4$ & $76.6 \pm 2.4$ \\ 
ViT \cite{vit} & $77.2 \pm 2.8$ & $76.1 \pm 2.7$ & $78.0 \pm 1.4$ & $76.6 \pm 2.4$ \\ 
DenseNet121 \cite{huang2017densely} & $77.0 \pm 1.0$ & $75.7 \pm 2.4$ & $77.4 \pm 1.3$ & $76.1 \pm 1.7$ \\ 
DenseNet201 \cite{huang2017densely} & $77.5 \pm 2.1$ & $76.5 \pm 2.1$ & $77.6 \pm 2.4$ & $76.8 \pm 2.2$ \\ 
SAR-JEPA \cite{li2024predicting} & $77.8 \pm 1.8$ & $75.9 \pm 3.7$ & $77.9 \pm 4.3$ & $76.8 \pm 2.3$ \\ 
SARATR-X \cite{li2025saratr} & $\underline{80.7 \pm 1.7}$ & $\underline{79.9 \pm 4.1}$ & $\underline{80.8 \pm 5.5}$ & $\underline{80.3 \pm 4.8}$ \\ 
\modelname{} & $\mathbf{82.9 \pm 0.7}$ & $\mathbf{80.8 \pm 1.5}$ & $\mathbf{81.9 \pm 1.4}$ & $\mathbf{81.3 \pm 1.4}$ \\ 
\bottomrule 
\end{tabular} 
\end{table}

\subsubsection{OpenSARShip Six-Class Scenario}

The OpenSARShip six-class scenario is considerably more challenging than the four-class scenario presented in Section~\ref{subsubsec:OpenSARShip_FourClass_Scenario}, as it introduces additional ship categories together with a more pronounced class imbalance. The considered classes include Bulk Carrier, Container Ship, Dredger, Fishing, Tanker, and Tug. Nevertheless, Table~\ref{tab:opensarship_6class} shows that the proposed method achieves the best performance across all evaluation metrics, obtaining an OA of 79.1\%, AA of 58.0\%, APr of 72.6\%, and AF1 of 59.8\%. 
In addition to the higher average performance, the proposed model also exhibits substantially lower standard deviations across repeated runs, indicating improved robustness and stability compared with the competing methods.
The improvement in AF1 is particularly notable, which provides a balanced assessment of precision and recall and is therefore more representative of performance under class imbalance. 
The results suggest that the proposed method not only improves the overall classification accuracy but also enhances class-wise discrimination, leading to a more balanced performance across the considered ship categories. 
Furthermore, the considerably higher average precision indicates that the proposed model produces more reliable predictions and is less prone to false-positive classifications.

\begin{figure*}[!t]
\centering
\includegraphics[width=0.9\textwidth]{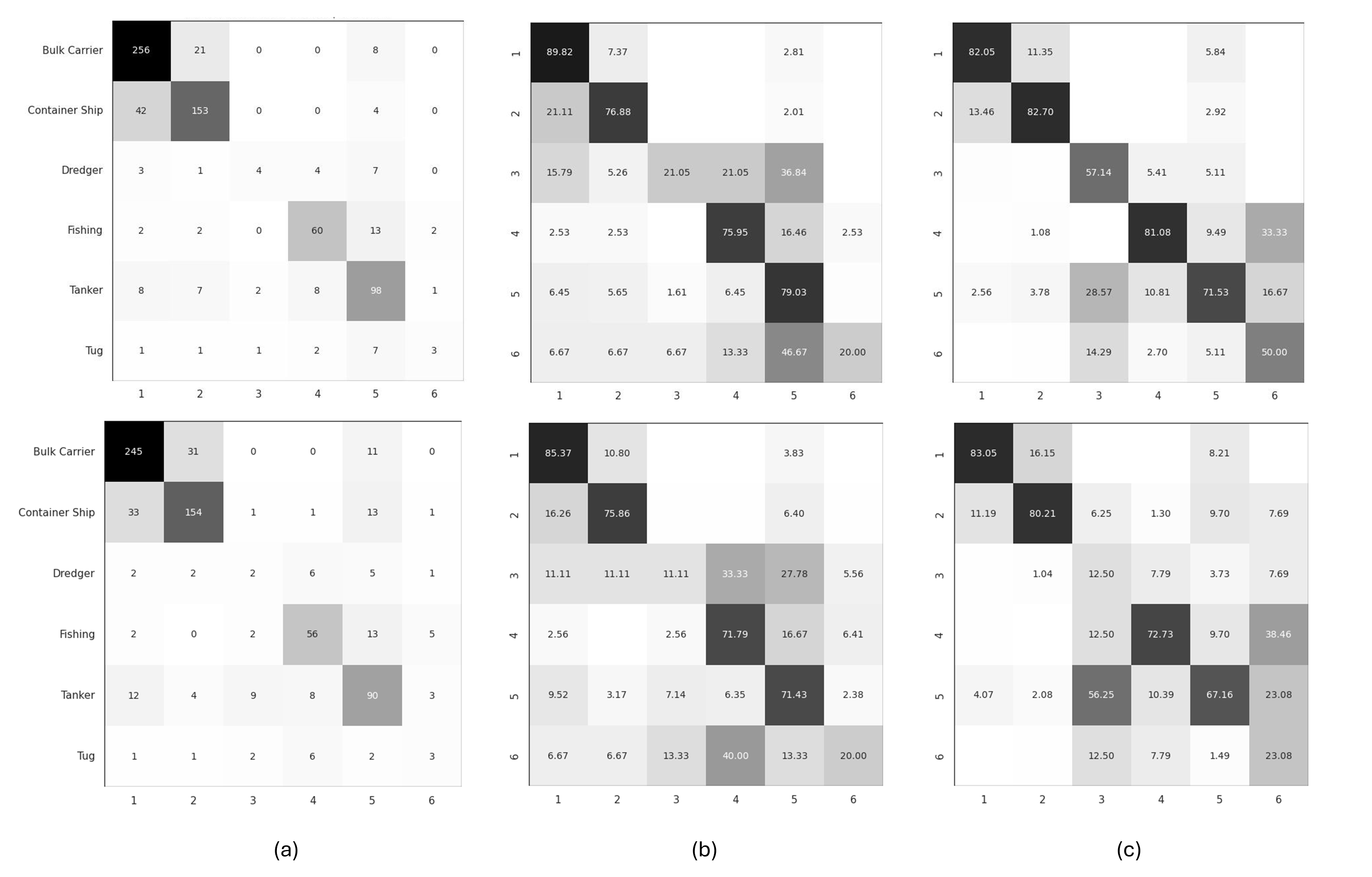}
\caption{Confusion matrices for the OpenSARShip six-class classification experiments. The first row shows the results of the proposed model, while the second row shows the results of the best-performing state-of-the-art model. (a) raw confusion matrix, (b) row-normalized confusion matrix (recall), and (c) column-normalized confusion matrix (precision).}\label{fig:confmat_opensarship6}
\end{figure*}

To further analyze the class-wise behavior, Fig.~\ref{fig:confmat_opensarship6} compares the confusion matrices of the proposed method and the best-performing state-of-the-art baseline. 
The row-normalized confusion matrices indicate that the proposed method achieves higher recall for most classes, demonstrating an improved ability to correctly identify samples from the true class.
The advantages of the proposed method become even more evident in the column-normalized confusion matrices. Stronger diagonal responses are observed for most classes, indicating higher precision and a reduced tendency toward false-positive predictions.
Overall, the improvement is particularly evident for the Dredger and Tug classes, which contain the fewest training samples and exhibit stronger inter-class confusion. The gains are especially pronounced in precision, indicating fewer false alarms for these challenging minority classes.

\begin{table}[!t] 
\footnotesize
\caption{Performance comparison on OpenSARShip six-class classification dataset (the best, and second-best performing results are shown in bold, and underlined, respectively).}
\label{tab:opensarship_6class}
\centering \footnotesize \setlength{\tabcolsep}{3pt} 
\begin{tabular}{lcccc} 
\toprule
Model & OA (\%) & AA (\%) & APr (\%) & AF1 (\%) \\ 
\midrule
ResNet50 \cite{resnet} & $68.0 \pm 2.9$ & $48.0 \pm 2.5$ & $48.6 \pm 4.8$ & $46.9 \pm 2.8$ \\
EffitNetV2-M \cite{efficientnetv2} & $67.1 \pm 1.5$ & $48.3 \pm 3.4$ & $46.3 \pm 1.9$ & $46.4 \pm 3.0$ \\
ViT \cite{vit} & $54.4 \pm 8.7$ & $34.1 \pm 10.1$ & $30.6 \pm 14.5$ & $30.4 \pm 12.3$ \\
DenseNet121 \cite{huang2017densely} & $73.4 \pm 2.0$ & $54.0 \pm 4.0$ & $55.9 \pm 8.1$ & $53.2 \pm 4.1$ \\
DenseNet201 \cite{huang2017densely} & $69.6 \pm 2.7$ & $49.1 \pm 3.0$ & $49.4 \pm 6.1$ & $48.3 \pm 3.0$ \\
SAR-JEPA \cite{li2024predicting} & $72.4 \pm 2.5$ & $54.0 \pm 5.6$ & $\underline{63.3 \pm 10.0}$ & $55.5 \pm 6.8$ \\
SARATR-X \cite{li2025saratr} & $\underline{75.7 \pm 1.3}$ & $\underline{55.9 \pm 4.1}$ & $56.5 \pm 9.9$ & $\underline{56.1 \pm 5.4}$ \\
\modelname{} & $\mathbf{79.1 \pm 0.9}$ & $\mathbf{58.0 \pm 1.5}$ & $\mathbf{72.6 \pm 0.7}$ & $\mathbf{59.8 \pm 1.9}$ \\
\bottomrule 
\end{tabular} 
\end{table}

\subsubsection{NASTaR}
The NASTaR dataset experiment considers a three-class ship classification scenario consisting of Cargo, Fishing, and Tanker vessels. The quantitative results presented in Table~\ref{tab:nastar} show that the proposed method achieves the best performance in terms of OA, APr, and AF1, obtaining 72.5\%, 73.6\%, and 69.4\%, respectively. 
In terms of AA, the proposed method achieves 68.1\%, which is the second-best result and remains highly competitive with the best-performing baseline. Compared with the strongest competing methods, the proposed approach improves OA by 1.7 percentage points (72.5\% vs. 70.8\%), APr by 2.4 percentage points (73.6\% vs. 71.2\%) both over EfficientNetV2-M, and AF1 by 0.5 percentage points over SARATR-X (69.4\% vs. 68.9\%). 
Although SARATR-X achieves a marginally higher AA (68.3\% vs. 68.1\%), the proposed method consistently provides stronger performance across the remaining evaluation metrics. Furthermore, the relatively small standard deviations demonstrate stable and robust behavior across repeated runs. The most notable gain is observed in average precision, indicating that the proposed method produces more reliable class assignments and is less prone to false-positive predictions.

\begin{figure*}[!t]
\centering
\includegraphics[width=0.9\textwidth]{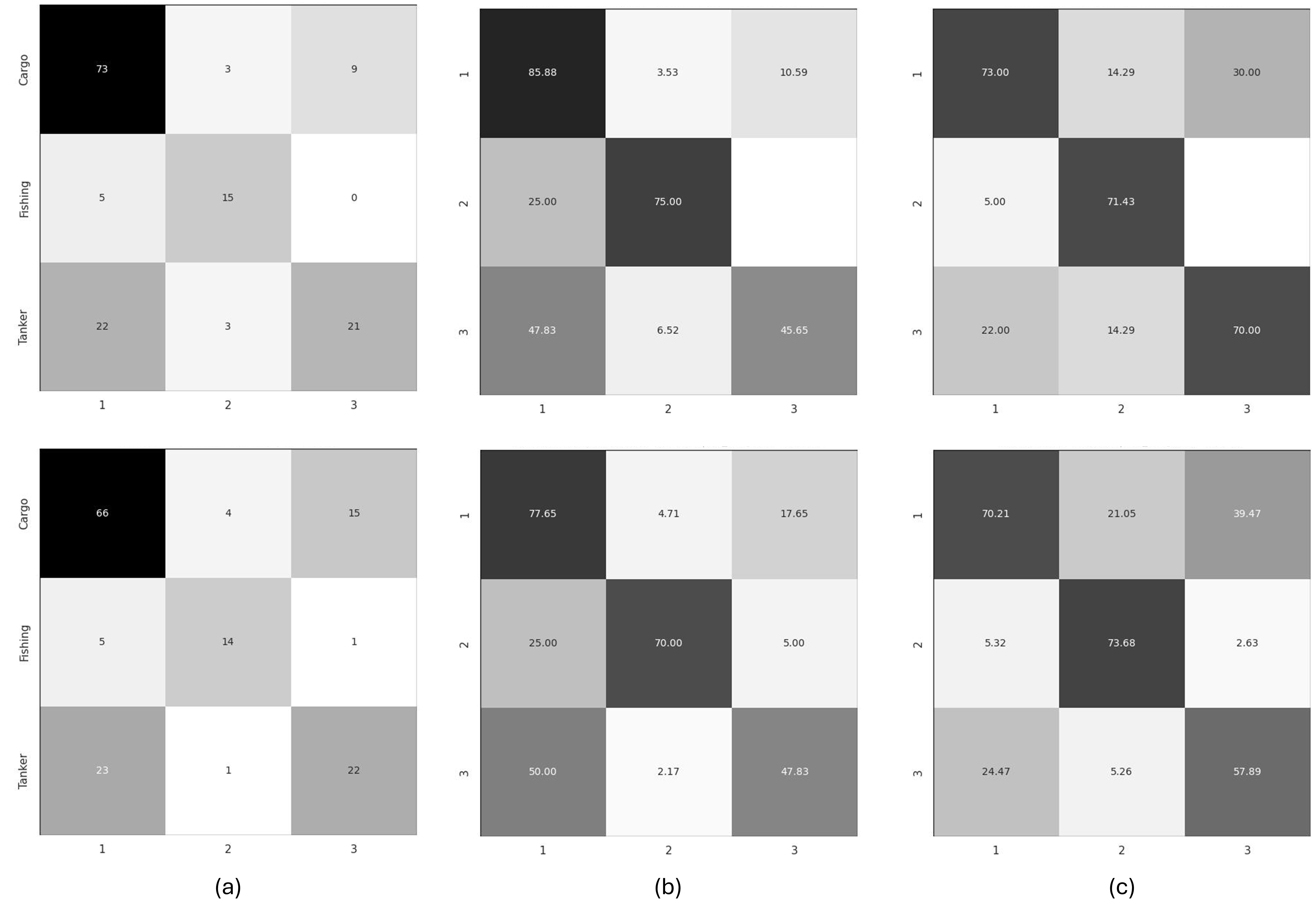}
\caption{Confusion matrices for the NASTaR classification experiments. The first row shows the results of the proposed model, while the second row shows the results of the best-performing state-of-the-art model. (a) raw confusion matrix, (b) row-normalized confusion matrix (recall), and (c) column-normalized confusion matrix (precision).}\label{fig:confmat_nastar}
\end{figure*}

To further investigate the class-wise performance, Fig.~\ref{fig:confmat_nastar} compares the confusion matrices of the proposed method and the best-performing state-of-the-art baseline. 
The row-normalized confusion matrices shown in Fig.~\ref{fig:confmat_nastar}(b) indicate that the proposed method achieves higher recall for the Cargo and Fishing classes. In particular, the recall of Cargo increases from approximately 77.7\% to 85.9\%, while the recall of Fishing improves from 70.0\% to 75.0\%. 
Tanker remains the most challenging category, exhibiting the close recall values between the proposed method and the baseline, respectively. The primary source of error for both methods arises from confusion between Cargo and Tanker vessels, as nearly half of the Tanker samples are incorrectly assigned to the Cargo class. Nevertheless, the proposed approach reduces several secondary misclassifications, leading to an overall improvement in class-wise recall.
The precision-oriented analysis shown in Fig.~\ref{fig:confmat_nastar}(c) provides further evidence of the effectiveness of the proposed method. Stronger diagonal responses are observed for all three classes, particularly for the Cargo and Tanker categories, where the Tanker precision improves substantially from approximately 57.9\% to 70.0\%. 
These results indicate that the proposed method generates fewer false positive predictions and produces more reliable vessel labels across all categories.
A closer inspection of the off-diagonal elements reveals that the proposed method noticeably reduces the mutual confusion between Cargo and Tanker vessels. While this remains the dominant classification challenge in the dataset, the lower off-diagonal responses and stronger diagonal elements demonstrate improved class separability. This behavior is particularly important because the reduction of false-positive predictions directly contributes to the higher average precision and F1-score reported in Table~\ref{tab:nastar}.

\begin{table}[!t] 
\footnotesize
\caption{Performance comparison on NASTaR dataset (the best, and second-best performing results are shown in bold, and underlined, respectively).}
\label{tab:nastar}
\centering \footnotesize \setlength{\tabcolsep}{3pt} 
\begin{tabular}{lcccc} 
\toprule 
Model & OA (\%) & AA (\%) & APr (\%) & AF1 (\%) \\ 
\midrule
ResNet50 \cite{resnet} & $67.0 \pm 2.9$ & $64.4 \pm 3.0$ & $66.2 \pm 3.2$ & $64.6 \pm 3.0$ \\
EffiNetV2-M \cite{efficientnetv2}  & $\underline{70.8 \pm 2.4}$ & $67.8 \pm 1.9$ & $\underline{71.2 \pm 3.4}$ & $68.5 \pm 2.4$ \\
ViT \cite{vit} & $66.8 \pm 3.6$ & $62.5 \pm 3.0$ & $68.0 \pm 4.0$ & $63.8 \pm 3.7$ \\
DenseNet121 \cite{huang2017densely} & $68.8 \pm 2.2$ & $66.4 \pm 2.4$ & $68.6 \pm 2.7$ & $66.8 \pm 2.4$ \\
DenseNet201 \cite{huang2017densely} & $70.2 \pm 2.7$ & $67.5 \pm 3.2$ & $70.3 \pm 3.2$ & $67.8 \pm 3.5$ \\
SAR-JEPA \cite{li2024predicting} & $67.4 \pm 3.2$ & $65.1 \pm 3.2$ & $66.9 \pm 1.7$ & $65.7 \pm 2.5$ \\
SARATR-X \cite{li2025saratr} & $69.6 \pm 0.6$ & $\mathbf{68.3 \pm 1.8}$ & $69.9 \pm 0.8$ & $\underline{68.9 \pm 1.1}$ \\
\modelname{} & $\mathbf{72.5 \pm 1.9}$ & $\underline{68.1 \pm 2.9}$ & $\mathbf{73.6 \pm 2.3}$ & $\mathbf{69.4 \pm 2.5}$ \\
\bottomrule
\end{tabular}
\end{table}

\subsubsection{FUSAR-Ship}
The FUSAR-Ship dataset consists of high-resolution SAR imagery containing five vessel categories, namely Bulk Carrier, Container Ship, Fishing, General Cargo, and Tanker in this experiment. Compared with OpenSARShip and NASTaR, the higher spatial resolution exposes richer structural details of vessels, but the dataset remains challenging due to the presence of visually similar ship types and the relatively limited number of samples in several categories.
The quantitative results presented in Table~\ref{tab:fusar} demonstrate that the proposed method achieves competitive performance across all evaluation metrics, obtaining an OA of 84.5\%, AA of 62.5\%, APr of 69.0\%, and AF1 of 64.3\%. Notably, the proposed approach achieves the highest average precision among all compared methods, outperforming the strongest competing baseline, SARATR-X, by 1.8 percentage points (69.0\% versus 67.2\%). The proposed model also attains the second-best OA, AA, and AF1, with performance differences of only 0.4, 0.6, and 0.3 percentage points, respectively, compared with the best-performing model. Furthermore, the lower standard deviations in OA and AF1 indicate a more stable behavior across repeated experimental runs.

The most significant advantage of the proposed method is reflected in average precision, which suggests that the model produces more reliable class assignments and is less prone to false-positive predictions. This trend is consistent with the observations from the OpenSARShip and NASTaR experiments, where the largest gains were also achieved in precision-oriented metrics.

\begin{figure*}[!t]
\centering
\includegraphics[width=0.9\textwidth]{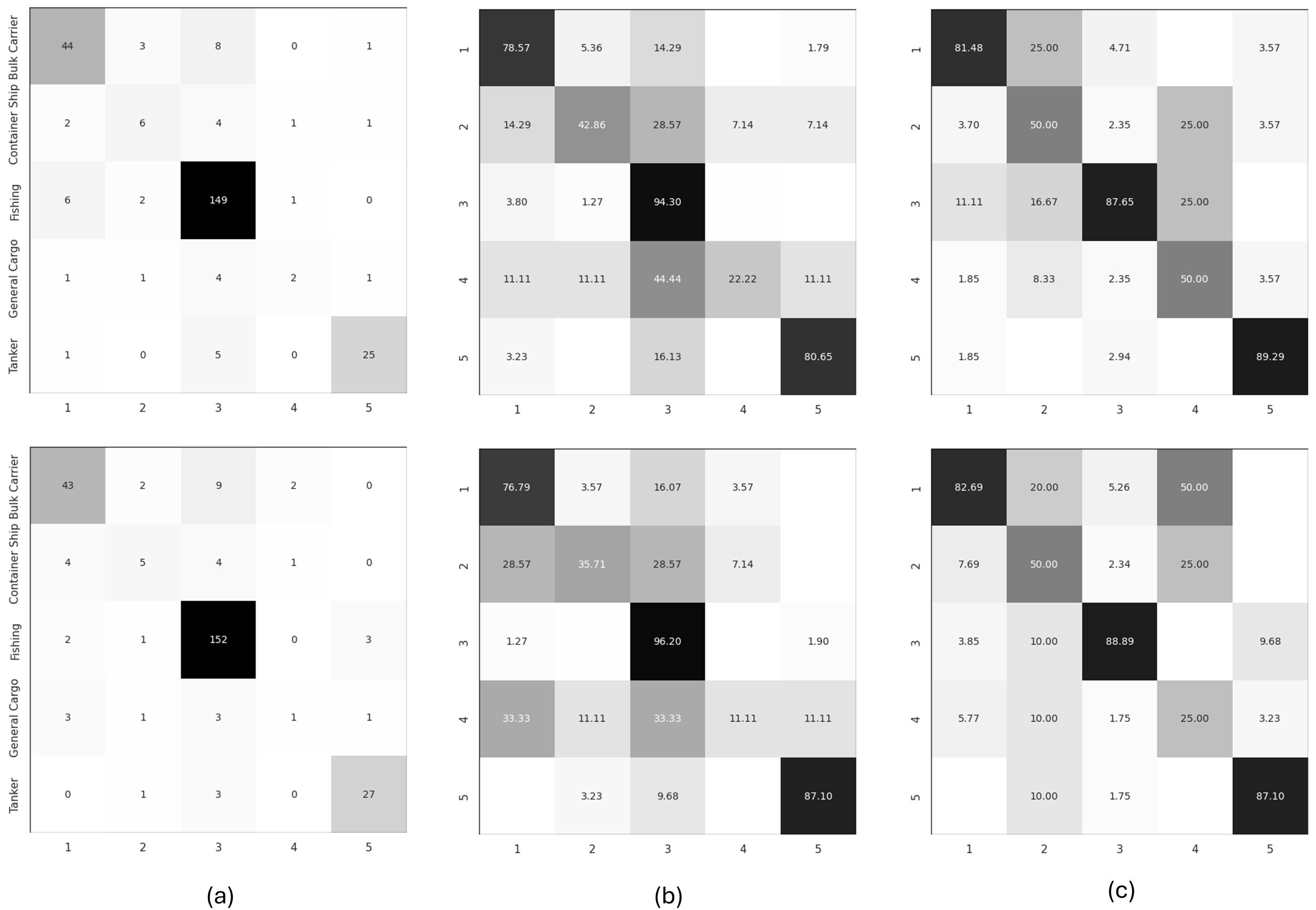}
\caption{Confusion matrices for the FUSAR classification experiments. The first row shows the results of the proposed model, while the second row shows the results of the best-performing state-of-the-art model. (a) raw confusion matrix, (b) row-normalized confusion matrix (recall), and (c) column-normalized confusion matrix (precision).}\label{fig:confmat_fusar}
\end{figure*}

To further investigate the class-wise behavior, Fig.~\ref{fig:confmat_fusar} compares the confusion matrices of the proposed method and SARATR-X, the best-performing state-of-the-art baseline.

From the row-normalized confusion matrices (Fig.~\ref{fig:confmat_fusar}(b)), it can be observed that the proposed method achieves higher recall for most categories. 
The most notable improvements are observed for Bulk Carrier and Container Ship, whose recall values increase from 69.6\% to 78.6\% and from 35.7\% to 42.9\%, respectively.
The precision-oriented confusion matrices in Fig.~\ref{fig:confmat_fusar}(c) reveal further advantages of the proposed model. The most notable improvement is observed for the General Cargo class, whose precision increases from 25.0\% to 50.0\%. Tanker precision also improves from 87.1\% to 89.3\%. However, the precision values of Bulk Carrier and Fishing change only marginally relative to the baseline. Overall, the stronger diagonal responses for General Cargo and Tanker indicate that the proposed method generates more reliable predictions and reduces false-positive assignments for these classes.

\begin{table}[!t] 
\footnotesize
\caption{Performance comparison on FUSAR-Ship dataset (the best, and second-best performing results are shown in bold, and underlined, respectively).}
\label{tab:fusar}
\centering \footnotesize \setlength{\tabcolsep}{3pt} 
\begin{tabular}{lcccc} 
\toprule 
Model & OA (\%) & AA (\%) & APr (\%) & AF1 (\%) \\ 
\midrule 
ResNet50 \cite{resnet} & $76.0 \pm 3.7$ & $49.1 \pm 5.3$ & $55.8 \pm 9.7$ & $49.9 \pm 5.3$ \\
EffiNetV2-M \cite{efficientnetv2} & $73.1 \pm 2.6$ & $43.3 \pm 3.0$ & $58.0 \pm 5.2$ & $46.7 \pm 2.8$ \\
ViT \cite{vit} & $62.3 \pm 4.6$ & $26.4 \pm 8.8$ & $19.2 \pm 10.2$ & $21.9 \pm 9.7$ \\
DenseNet121 \cite{huang2017densely} & $79.4 \pm 3.1$ & $56.5 \pm 3.2$ & $58.5 \pm 4.2$ & $56.4 \pm 2.6$ \\
DenseNet201 \cite{huang2017densely} & $81.2 \pm 1.9$ & $54.4 \pm 2.3$ & $58.5 \pm 2.9$ & $54.9 \pm 1.8$ \\
SAR-JEPA \cite{li2024predicting} & $82.4 \pm 1.3$ & $56.1 \pm 3.0$ & $57.4 \pm 5.3$ & $56.3 \pm 3.9$ \\
SARATR-X \cite{li2025saratr} & $\mathbf{84.9 \pm 1.3}$ & $\mathbf{63.1 \pm 3.9}$ & $\underline{67.2 \pm 4.8}$ & $\mathbf{64.6 \pm 4.2}$ \\
\modelname{} & $\underline{84.5 \pm 1.1}$ & $\underline{62.5 \pm 2.5}$ & $\mathbf{69.0 \pm 4.0}$ & $\underline{64.3 \pm 2.7}$ \\
\bottomrule 
\end{tabular} 
\end{table}

\subsection{Effect of Training Sample Size}

To assess the robustness of the proposed model under limited training data, experiments were conducted on OpenSARShip using different training sample percentages. The proposed method was compared with baseline models under the same data split conditions. This experiment is particularly important because SAR ship classification datasets often contain limited labelled samples especially for minority classes.

\begin{figure}[!t]
\centering
\includegraphics[width=0.5\textwidth]{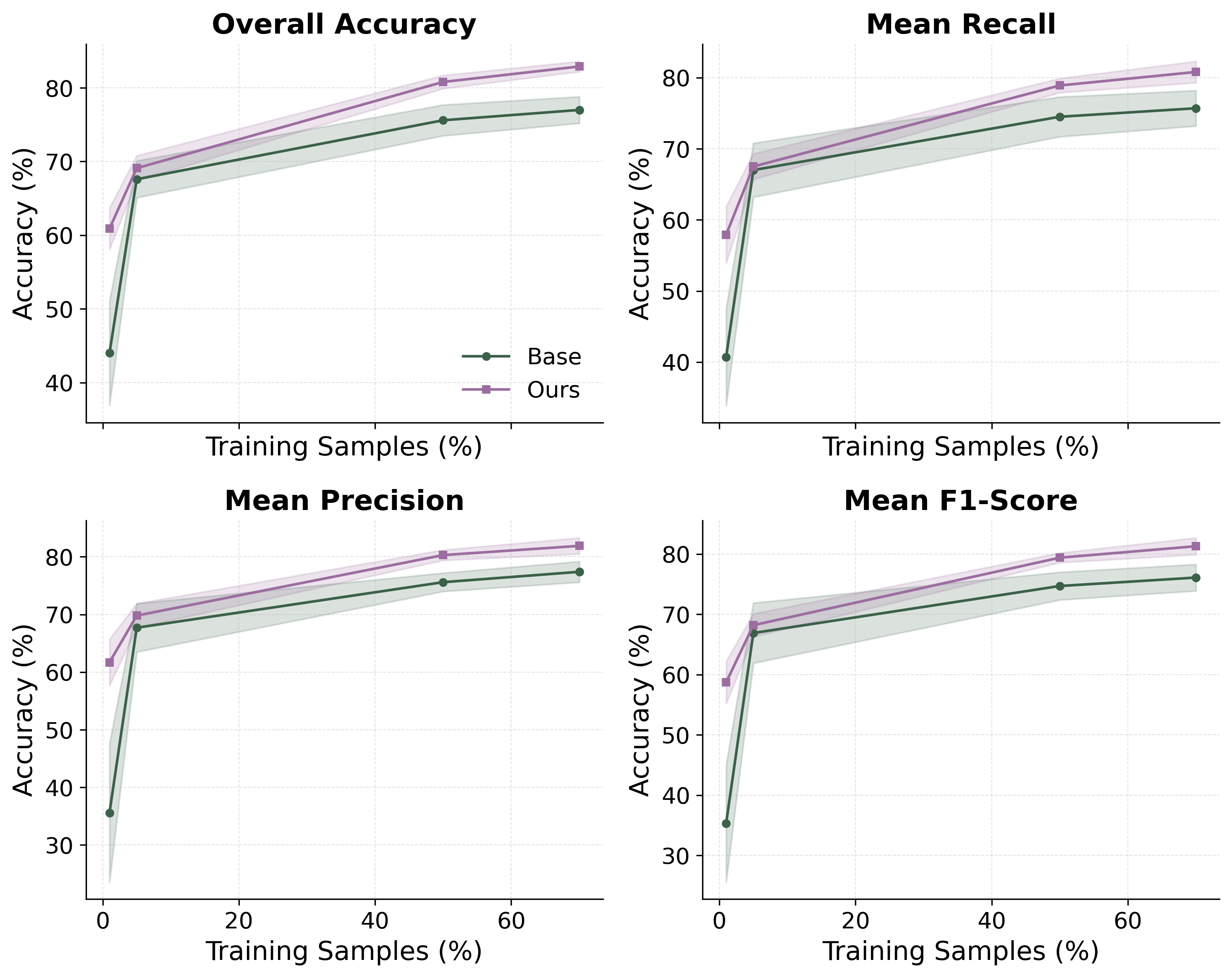}
\caption{Performance under different training sample percentages on OpenSARShip.}\label{fig:training_percentage}
\end{figure}

Fig.~\ref{fig:training_percentage} illustrates the impact of different training sample percentages on model performance. The proposed method consistently outperforms the baseline across all evaluation metrics and training settings. The most significant improvement is observed when only 1\% of the training samples are available, where the proposed model improves OA from 44.0\% to 60.9\%, AA from 40.7\% to 57.9\%, APr from 35.6\% to 61.7\%, and AF1 from 35.3\% to 58.7\%. These substantial gains demonstrate the effectiveness of the proposed approach in data-scarce scenarios.

In addition to achieving higher accuracies, the proposed method exhibits considerably lower standard deviations, particularly at 1\% training data. For example, the standard deviation of APr decreases from 12.0 to 4.0 and that of AF1 decreases from 9.7 to 3.5, indicating improved robustness and reduced sensitivity to training sample selection. As the amount of training data increases, the performance of both methods improves; however, the proposed model consistently maintains superior results, achieving OA values of 80.8\% and 82.9\% at 50\% and 70\% training samples, respectively. These results suggest that the proposed framework not only benefits from larger training sets but also generalizes effectively under limited-data conditions.

\subsection{Ablation Studies}

Ablation studies were conducted on the OpenSARShip four-class scenario using 50\% training, 10\% validation, and the remaining samples for testing. The aim was to evaluate the influence of the multi-scale feature extraction depth, the Directional Sparse Convolution stem, and the proposed imbalance-aware metric-learning loss.

\begin{table}[!t] 
\footnotesize
\caption{Ablation study results on the OpenSARShip four-class scenario.}
\label{tab:ablation_opensarship}
\centering \footnotesize \setlength{\tabcolsep}{3pt} 
\begin{tabular}{lcccc}
\toprule
Configuration & OA (\%) & AA (\%) & APr (\%) & AF1 (\%) \\ 
\midrule
MSFE1: 64 & $77.0 \pm 1.4$ & $75.3 \pm 1.4$ & $75.3 \pm 1.7$ & $75.2 \pm 1.5$ \\
MSFE2: 64, 128 & $78.8 \pm 0.9$ & $77.3 \pm 1.0$ & $77.5 \pm 1.6$ & $77.3 \pm 1.0$ \\
MSFE3: 64, 128, 256 & $\mathbf{80.8 \pm 0.9}$ & $\mathbf{78.9 \pm 1.0}$ & $\mathbf{80.3 \pm 0.9}$ & $\mathbf{79.4 \pm 0.8}$ \\
MSFE4: 64, 128, 256, 512 & $80.2 \pm 0.6$ & $78.5 \pm 0.6$ & $79.7 \pm 0.7$ & $79.0 \pm 0.3$ \\
No Directional: 64, 128, 256 & $80.1 \pm 1.5$ & $78.2 \pm 1.4$ & $79.1 \pm 1.4$ & $78.5 \pm 1.4$ \\
Cross-Entropy: 64, 128, 256 & $79.7 \pm 1.0$ & $77.5 \pm 1.2$ & $79.7 \pm 0.9$ & $78.4 \pm 0.9$ \\
\bottomrule 
\end{tabular} 
\end{table}

As shown in Table~\ref{tab:ablation_opensarship}, increasing the depth of the multi-scale feature extractor improves performance up to the three-block configuration. The MSFE3 setting achieves the best results across all four metrics. Adding a fourth MSFE block does not further improve performance, suggesting that the three-block design provides a better balance between representation capacity and model complexity.

Excluding the directional module reduces performance compared with the full model, confirming the contribution of the Directional Sparse Convolution stem. Similarly, replacing the proposed loss function with standard cross-entropy decreases the performance, indicating that the imbalance-aware metric-learning objective improves embedding separability beyond conventional supervised classification.

\subsection{Computation and Efficiency}

\begin{figure}[ht]
\centering
\includegraphics[width=0.5\textwidth]{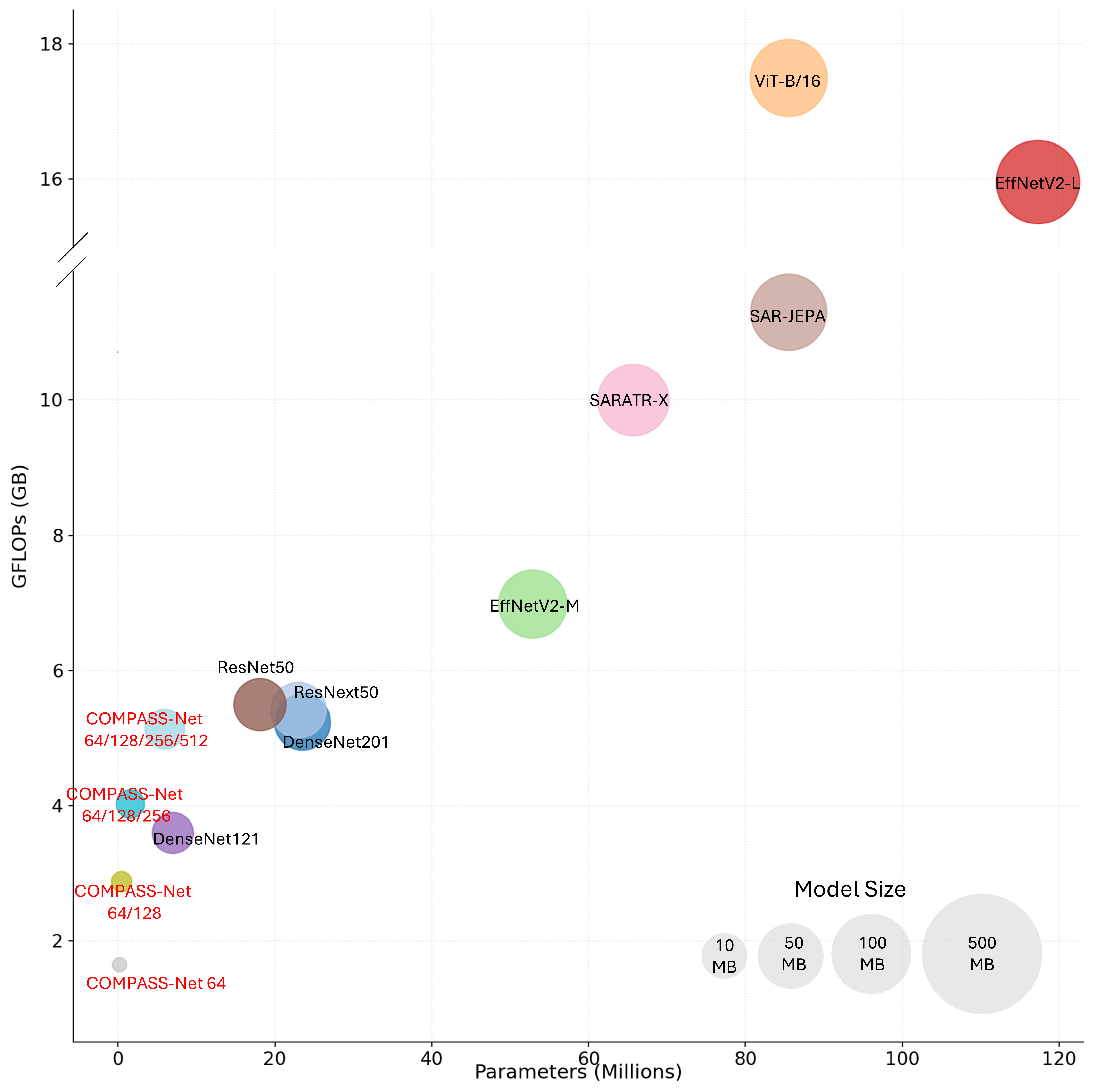}
\caption{Model complexity comparison with the state-of-the-art. }\label{fig:model_complexity}
\end{figure}

Fig ~\ref{fig:model_complexity} compares the complexity characteristics of the proposed multi-scale directional convolutional architectures against widely adopted convolutional and transformer-based models. 
The proposed networks employ a hierarchical multi-scale feature extraction strategy, in which convolutional layers are progressively stacked to capture image structures at different receptive fields. To investigate the scalability of the architecture, four variants were evaluated, consisting of one, two, three, and four feature extraction stages with filter configurations of 64, 64/128, 64/128/256, and 64/128/256/512 channels, respectively.
For a fair comparison, all FLOP measurements were computed using an input image size of $256\times256$ pixels and a batch size of one.
The results demonstrate that all the experimented variants occupy a highly favorable region of the complexity space, located toward the lower-left corner of the plot. This indicates that the networks achieve considerably lower parameter counts and lower computational requirements compared with state-of-the-art deep convolutional and transformer architectures. 
The smallest variant, containing a single 64-channel directional feature extraction layer, requires only 0.104 million parameters and 1.65 GFLOPs, corresponding to a model size of only 0.41 MB. Despite its compact design, this lightweight configuration achieved performance comparable to larger state-of-the-art architectures. 
As the number of multi-scale feature extraction stages increases, the representational capacity of the network improves further, with the deeper variants achieving state-of-the-art performance while maintaining a lower computational footprint than competing models.

A particularly noteworthy observation is the scalability of the proposed architecture. Increasing the model capacity from the smallest to the largest variant results in a parameter growth from only 0.104 million to 5.91 million parameters, while the computational complexity increases from 1.65 to 5.14 GFLOPs. Even the largest proposed configuration remains significantly smaller than DenseNet201 (18 million parameters), ResNet50 (23 million parameters), EfficientNetV2-M (53 million parameters), SARATR-X (65M parameters and approximately 10 GFLOPs), and base-ViT with $16 \times 16$ patches (85 million parameters). This behavior demonstrates that the developed module can be effectively scaled to increase representational power without incurring the substantial parameter growth typically associated with conventional convolutional architectures.

Overall, the results indicate that the proposed multi-scale directional feature extraction framework offers a balance between representational capacity and computational efficiency. The ability to maintain low parameter counts, low FLOPs, and small storage requirements across multiple architectural scales demonstrates both the efficiency and scalability of the proposed module.

\begin{figure}[t]
\centering
\includegraphics[width=0.5\textwidth]{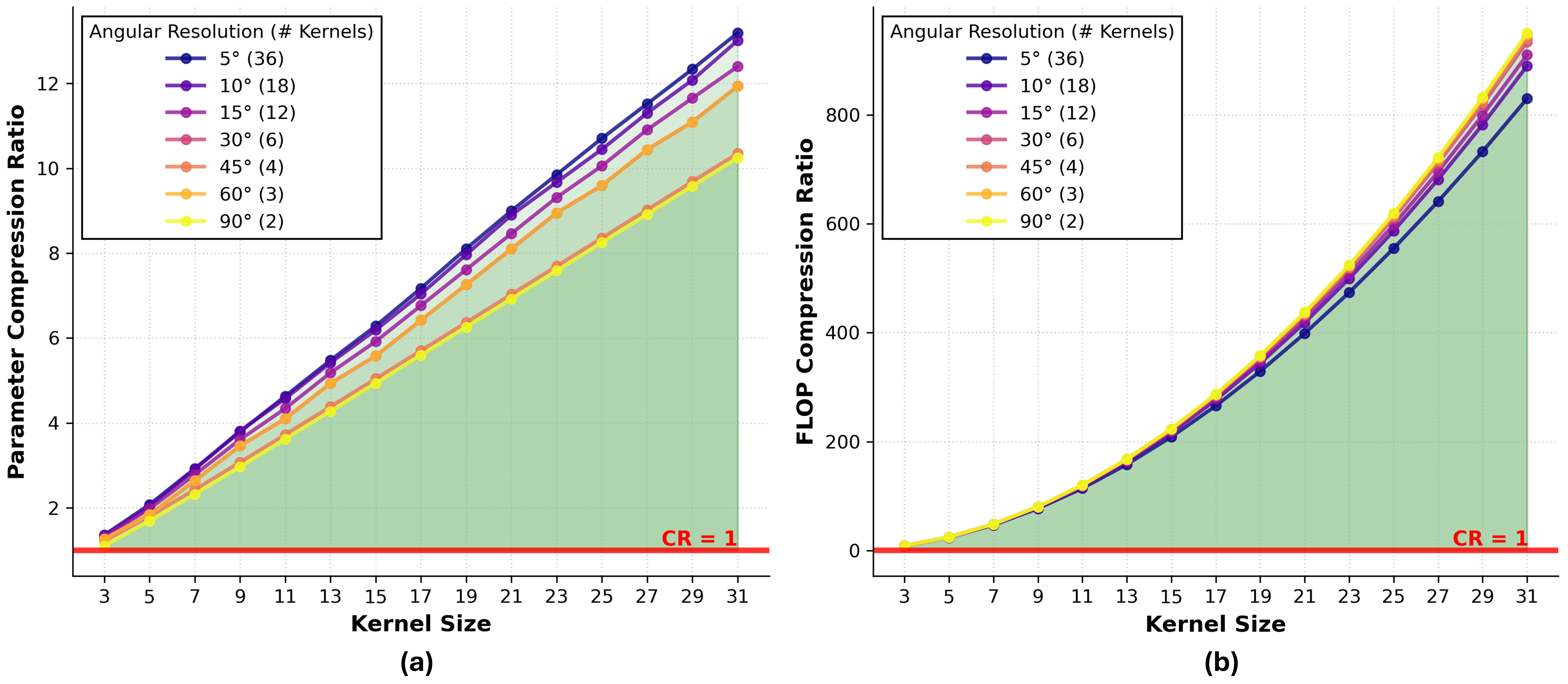}
\caption{Compression ratio of the proposed directional convolution relative to a conventional 2D convolution as a function of kernel size for different angular resolutions: a) Parameter compression ratio. b) FLOP compression ratio. }\label{fig:directional_complexity}
\end{figure}

To further evaluate the computational efficiency of the proposed directional convolution module, experiments were performed using kernel sizes ranging from \(3\times3\) to \(31\times31\) and angular intervals between \(5^\circ\) and \(90^\circ\), corresponding to 36 and 2 directional filter masks, respectively. 
Fig.~\ref{fig:directional_complexity} illustrates the compression efficiency of the proposed directional convolution relative to a conventional 2D convolution across different kernel sizes and angular resolutions. The compression ratio (CR) is defined as the ratio of the complexity of the standard 2D convolution to that of the directional convolution, such that values greater than one indicate a more efficient directional representation.
Across all investigated configurations, the proposed module consistently achieved compression relative to an equivalent conventional 2D convolution in both the number of trainable parameters and FLOPs, demonstrating its computational advantage over standard convolutional filtering. As shown in Fig.~\ref{fig:directional_complexity}-a, the parameter compression ratio increases approximately linearly with kernel size, indicating progressively larger parameter savings as the receptive field grows. A similar but more pronounced trend is observed for FLOP compression (Fig.~\ref{fig:directional_complexity}-b), where the computational savings increase rapidly with kernel size. 
These results highlight a key characteristic of the proposed formulation: its efficiency advantage becomes increasingly significant for large kernels, which are traditionally associated with substantial parameter and computational overhead in conventional convolutional layers. 
Consequently, the directional convolution module enables the incorporation of large receptive fields at a fraction of the parameter and computational cost required by standard convolutions. Furthermore, the module exhibits excellent scalability, as its compression benefits are consistently maintained across the entire range of investigated kernel sizes and angular resolutions, allowing the complexity-performance trade-off to be adjusted through the number of directional masks without sacrificing computational efficiency. The fact that compression is achieved in all tested configurations further demonstrates the robustness and general applicability of the proposed design, making it suitable for both resource-constrained and large-scale deep learning architectures.

\section{Conclusion}
\label{sec:Conclusion}
This paper presented \modelname{}, a class-frequency-aware metric learning framework for SAR ship-type classification under imbalanced training conditions. 
The proposed method was designed to address several persistent challenges in SAR ship recognition, including limited labeled data, class imbalance, large intra-class variability, fine-grained inter-class similarity, and variations in target scale, orientation, and scattering structure. 
To this end, the framework integrates dual-view input preparation, directional sparse convolution, hierarchical multi-scale feature extraction, dual-branch feature fusion, and imbalance-aware metric learning within a unified architecture. 
The proposed dual-view preparation strategy allows the model to jointly exploit target-centered structural details and pooled contextual information. 
The directional sparse convolutional stem improves the extraction of orientation-sensitive ship features by explicitly modeling elongated and direction-dependent scattering responses.
The customized hierarchical multi-scale feature extraction module further enhances representation learning by combining dilated convolutional branches for multi-scale contextual modeling with deformable convolution for geometric adaptability. 
To address the effect of imbalanced class distributions, a class-frequency-aware metric learning objective was introduced. The proposed loss combines margin-based classification, class-frequency weighted prototype compactness, and class-pair frequency-weighted margin separation. By explicitly regularizing the geometry of the embedding space, the loss encourages compact class-specific clusters while enforcing stronger separation for class pairs involving rare ship categories. This improves minority-class discrimination and reduces the tendency of majority classes to dominate the learned representation.
Comprehensive experiments on OpenSARShip, FUSAR-Ship, and NASTaR demonstrated the effectiveness and robustness of the proposed framework across different SAR sensors, resolutions, and acquisition domains. The ablation studies confirmed the contribution of the main architectural and loss-function components, including the directional sparse convolution module, the hierarchical multi-scale feature extractor, and the proposed metric-learning objective. 
Comparative experiments further showed that \modelname{} consistently achieves superior performance over advanced deep-learning models in both overall accuracy and class-balanced metrics. These results indicate that the proposed framework is well suited for practical SAR ship type classification, particularly in scenarios involving limited labeled data and imbalanced class distributions. 
The proposed architecture also demonstrates advantages in terms of model efficiency. In particular, the directional sparse convolution module requires substantially fewer parameters than conventional dense convolutional kernels while still capturing orientation-aware ship structures. This makes it suitable for efficient feature extraction and suggests strong scalability to larger kernels, where wider receptive fields can be obtained without a proportional increase in parameter count.

\bibliographystyle{IEEEtran} 
\bibliography{cas-refs.bib}

\end{document}